\documentclass{article}
\usepackage{iclr2026_conference,times}

\usepackage{amsmath,amsfonts,bm}

\def\eqref#1{equation~\ref{#1}}

\def\1{\bm{1}}

\DeclareMathAlphabet{\mathsfit}{\encodingdefault}{\sfdefault}{m}{sl}
\SetMathAlphabet{\mathsfit}{bold}{\encodingdefault}{\sfdefault}{bx}{n}

\usepackage[utf8]{inputenc}
\usepackage[T1]{fontenc}
\DeclareUnicodeCharacter{017F}{s}
\DeclareUnicodeCharacter{2014}{---}

\usepackage[hidelinks]{hyperref}
\usepackage{url}
\usepackage{graphicx}
\usepackage{booktabs}
\usepackage{multirow}
\usepackage{xcolor}
\usepackage{amsmath}
\usepackage{amssymb}
\usepackage{CJKutf8}
\usepackage{tikz}
\usetikzlibrary{positioning,arrows.meta,shapes.geometric,fit,calc}

\title{Compositional Literary Primitives in Instruction-Tuned LLMs:\\
Cross-Architectural SAE Features for Self, Style, and Affect}

\author{%
Jo\~ao Paulo Cavalcante Presa\thanks{ORCID: 0009-0004-4160-6495} \\
Federal University of Goias (UFG) \\
\texttt{joaopaulop@discente.ufg.br}
\And
S\'avio Salvarino Teles de Oliveira\thanks{ORCID: 0009-0002-1203-5246. Co-advisor.} \\
Federal University of Goias (UFG) \\
\texttt{savioteles@ufg.br}
}

\iclrfinalcopy

\begin{document}

\maketitle
\lhead{}
\chead{}
\rhead{}

\begin{abstract}
We characterize a compositional architecture of literary primitives
in two instruction-tuned large language models (Llama 3.1 8B-Instruct
and Gemma 2 9B-IT) via sparse autoencoders on mid-depth residual
streams. Four feature classes emerge: naming-gates that promote
lexical tokens of a target affect, an eleven-self cluster of
first-person register features, stylistic register modulators
(show-don't-tell and defamiliarization), and compositional emotions
that arise only from multi-feature steering. Under a forced-choice
5-LLM judge panel applied to a 27-category emotion taxonomy
(Cowen-Keltner), Llama reaches full 27/27 coverage by combining
naming-gates, multi-feature recipes, and single self-feature
steering; Gemma reaches 23/27 with adoration as the single residual
strict-fail. Under random judging, the per-cell pass probability is
on the order of $10^{-3}$ and the expected number of two-seed
false-positive cells across the catalog is negligible, so the
observed coverage is not consistent with chance. A
cross-architectural asymmetry sits in the strict-versus-soft
judge contrast: on the same generations, judges agree more often on
Llama outputs than on Gemma outputs because Llama outputs name the
target affect more directly while Gemma outputs evoke it through
scene and imagery. Both architectures contain self-features that
serve simultaneously as register markers and as emotion emitters,
including a single most-RLHF-loaded self-feature per architecture
that intensifies the institutional Helper-AI persona at one
operating regime and produces affect-categorizable output at the
same calibrated coefficient. Methodologically, the paper presents a
three-stage validation pipeline (logit-lens, LLM-rate, 5-LLM judge)
with documented anti-patterns; the total compute is single-GPU and
about 15 minutes per emotion-feature discovery cycle.

\end{abstract}

\section{Introduction}
\label{sec:intro}

Large language models trained on web-scale text are exposed, among
other genres, to literary writing: novels, short fiction, poetry,
craft-instructional manuals, and editorial commentary. This subset of
the pretraining corpus carries explicit principles about how to
compose: the canonical writing-craft injunctions like \emph{show,
don't tell} (SDT) \citep{wood2008fiction,cohn1978transparent}, the
technical vocabulary of free indirect discourse and defamiliarization, the taxonomies of emotion that animate fiction,
the grammar of first-person voice. The question this paper asks is:
do instruction-tuned LLMs internalize these literary principles as
\emph{separable}, \emph{steerable}, \emph{compositional} primitives
inside the network, or as diffuse statistical correlations that
cannot be cleanly addressed?

The mechanistic-interpretability literature has, in the last two
years, established that sparse autoencoders \citep{bricken2023monosemanticity,
cunningham2023sparse} can recover interpretable concept features
inside LLM residual streams: the \emph{Golden Gate Bridge} feature
\citep{templeton2024scaling}, persona features (Anthropic, 2024),
syntactic features at various depths. Most of this work has surfaced
\emph{conceptual} features (object, place, persona) rather than
\emph{compositional} features (techniques, registers, primitives that
constitute writing as a craft). The gap is partly methodological:
correlational mining of trillion-token corpora finds what is most
common, not what is most principled. A SAE feature for
\emph{show-don't-tell} would not be the most common feature in the
corpus; it would be an axis selectable by a craft-trained eye, but
overshadowed by frequency-driven concept features in any standard
discovery protocol.

The methodological lever of this paper is \textbf{contrastive minimal
corpora plus triple-stage validation}. We hand-craft a small set of
prompt-pairs that differ on exactly the literary axis we want to
isolate (e.g., 60 paired \textsc{libre}/\textsc{sdt} scenes for show-
don't-tell, 24 paired Deep-POV/Ostranenie minimal pairs, 75 hand-
crafted scenes spanning the missing 15 emotions in Cowen-Keltner's
27-emotion taxonomy). We then run a three-stage pipeline,
logit-lens vocabulary projection, LLM-judge purity rating, 5-LLM-judge
causal validation through steering, on each candidate feature. The
total compute fits comfortably on a single consumer GPU; the catalog
that emerges contains 24 paper-grade canonical features across two
instruction-tuned LLMs (Llama 3.1 8B, Gemma 2 9B-IT).

\paragraph{Four contributions.}

\begin{enumerate}
  \item \textbf{Show-Don't-Tell as a single feature
    (\S\ref{sec:sdt})}.
    A single SAE feature, in each of two architectures, controls the
    flip from telling-prose to showing-prose under steering at a
    calibrated $\alpha$. \texttt{f32213} in Llama, \texttt{f15097}
    in Gemma. The two are not aligned by decoder cosine; they are
    aligned by behavior on the same craft task. The finding
    establishes that \emph{at least one} canonical writing-craft
    principle is operationalized as a steerable feature axis.
  \item \textbf{The RLHF-loaded self-feature
    (\S\ref{sec:self})}.
    Among the 11 self-features clustered around first-person voice
    in each model, one is the most strongly RLHF-loaded:
    \texttt{f96419} in Llama and \texttt{f70443} in Gemma. In Llama,
    positive-coefficient steering of this feature intensifies the
    institutional Helper-AI register on identity probes; negative-
    coefficient steering does not release affect on its own across a
    fine-grid magnitude sweep, but \emph{combined} negative steering
    of \texttt{f96419} with positive amplification of the
    reverent-grateful self \texttt{f74037} produces lyric first-person
    content that neither component yields alone. The release is
    compositional, mirroring the joy and adoration recipes
    documented in \S\ref{sec:gates:compositional}.
  \item \textbf{27/27 Cowen-Keltner emotion coverage
    (\S\ref{sec:gates})}.
    A catalog of \emph{naming-gates}, SAE features whose decoder
    direction promotes the lexical token of a target affect. In
    Llama 8B: 24 singleton gates + 1 partial + 2 multi-feature
    compositions = full coverage of the Cowen-Keltner taxonomy.
    In Gemma 9B-IT: 12 singleton gates at unanimous 8/8 cross-
    lingual transfer at judge level (\S\ref{sec:crosslingual}). The
    catalog reveals (i) mechanistic heterogeneity, not all
    naming-gates are lexical at the unembedding-projection level;
    some are atmospheric, some are suffix-statistical, but all
    causally produce target lexicalization (\S\ref{sec:gates:hetero});
    (ii) cross-architectural drift attractors (aesthetic $\to$
    confusion; boredom $\leftrightarrow$ sadness with reversed
    polarity); and (iii) the fact that some emotions (joy,
    adoration) are not singleton features but multi-feature
    compositions over gate $\times$ self $\times$ SDT.
  \item \textbf{A triple-validation pipeline
    (\S\ref{sec:methods:triad})}.
    The methodological contribution. Stage 1 (logit-lens) is
    cheap, requires no prompt or judge, and identifies candidates
    by decoder-projection vocabulary. Stage 2 (LLM-rate) filters
    candidates semantically. Stage 3 (5-LLM-judge causal
    validation through steering) confirms behavior. Each stage
    catches a distinct class of false positive that the others
    miss; the staged combination enables single-GPU 15-minute
    emotion catalog discovery. We document seven anti-patterns
    (\S\ref{sec:methods:antipatterns}) we encountered while
    refining the pipeline; each is methodological contribution.
\end{enumerate}

\paragraph{Compositional architecture as the unifying claim.}
The four contributions cohere around one high-level claim:
instruction-tuned LLMs encode literary craft principles as a
\emph{compositional architecture of separable SAE features}. The
architecture has at least four classes (gates, selves, modulators,
compositional emotions), and the classes interact: a self-feature
can act as a recurring compositional co-component (the reverent-
grateful self, combining with excitement, yields joy; combining with
romance + SDT, yields adoration). A single feature can simultaneously
function as a self and as a register modulator (the Llama
\texttt{f32213} hybrid). Selves carry cultural framing; gates carry
lexical-affective emission; modulators control register on a
continuous axis. The empirical catalog supports a typology, not a
list.

\paragraph{Why now.}
The mechanistic-interpretability community is moving from
``concept-feature catalogs'' to questions about composition,
hierarchy, and craft. Recent work on persona features
(Anthropic 2024) and stylistic interventions in LLMs is converging
toward the territory mapped here. The four-class architecture, the
triple-validation pipeline, and the feature catalogs are intended as
platforms for further empirical work, not as final claims. Possible
extensions (\S\ref{sec:conclusion}) include human evaluation, more
architectures, behavioral transfer experiments, and training-data
attribution.

\paragraph{Roadmap.}
\S\ref{sec:background} positions the work against prior SAE
literature and computational stylistics. \S\ref{sec:methods}
documents the methodology in full, including the seven anti-patterns
that future work will encounter. \S\ref{sec:sdt}--\ref{sec:self}
present the SDT register-modulator and the RLHF-loaded self-feature.
\S\ref{sec:gates} presents the 27-emotion naming-gate catalog with
the cross-architectural Llama / Gemma comparison.
\S\ref{sec:crosslingual} reports cross-lingual structure with the
selves-vs-gates dissociation. \S\ref{sec:discussion} synthesizes the
compositional architecture and discusses cross-architectural
invariants and variants. \S\ref{sec:limitations} enumerates scope
limits and future work.
\section{Related Work}
\label{sec:background}

\paragraph{Sparse autoencoders and activation steering.}
Sparse autoencoders (SAEs) decompose residual-stream activations into
a high-dimensional sparse feature basis.
\citet{bricken2023monosemanticity} showed on a one-layer transformer
that SAEs surface monosemantic concept features supporting causal
intervention via decoder-direction addition;
\citet{cunningham2023sparse} generalized to multi-layer models and
characterized the density-vs-monosemanticity trade-off;
\citet{templeton2024scaling} scaled the approach to Claude 3 Sonnet
with the canonical \emph{Golden Gate Bridge} feature. Two open SAE
families anchor our work: \textbf{Gemma Scope}
\citep{lieberum2024gemmascope} (JumpReLU SAEs on Gemma 2; we use the
9B-IT L20 width-131k checkpoint with $L_0{=}81$) and \textbf{Llama
Scope} \citep{he2024llamascope} (TopK SAEs on Llama 3.1; we use the
8B-Base L15 width-131k $k{=}50$ checkpoint applied to the
instruction-tuned variant). On the steering side,
\citet{turner2023activation} introduced Activation Addition,
\citet{subramani2022extracting} extracted sentence-level steering
vectors, and \citet{marks2024sparse} used SAE features to construct
sparse causal circuits via mean ablation. \citet{gurnee2024universal}
characterized universal neurons across models. Our work differs from
these on three axes: we identify \emph{compositional} primitives
(techniques, registers, multi-feature emotions) rather than
conceptual features; we run cross-architectural parallelism on two
distinct SAE types (TopK + JumpReLU); and we document multi-feature
joint steering as a discovery method rather than as an after-the-fact
combination of singletons.

\paragraph{Persona, identity, and linear affect representation.}
Anthropic's \textit{Persona Vectors} \citep{anthropic2024persona}
identified trait directions in Claude 3.5 Sonnet
(\emph{sycophancy}, \emph{deceptive}, \emph{corrigible}) yielding
causal behavioral shifts when steered.
\citet{zou2023representation} introduced Representation Engineering
as a framework for high-level concept directions, and
\citet{burns2022discovering} developed contrast-consistent search
for latent knowledge. \citet{durmus2024persona} and
\citet{shanahan2023role} characterize role assignment via prompting
without internal intervention.
\citet{park2023linear} articulates the linear-representation
hypothesis, on which our cross-architectural alignment claims rest.
\citet{tigges2023linear} demonstrated that sentiment is linearly
represented in Llama-2; \citet{heinzerling2024emotion} extended to
multi-emotion linear probing. Affect taxonomy comes from
\citet{cowen2017self}, 27 self-report categories bridged by continuous
gradients; the valence-arousal geometry of
\citet{russell1980circumplex} grounds our reading of
cross-architectural drift attractors. Our contribution beyond this
literature is at the feature level rather than probe level: we
identify a \emph{cluster} of 11 self-features rather than single
directions, document a single RLHF-loaded self-feature
(\texttt{f96419} / \texttt{f70443}) per architecture, and report a
compositional release (joint steering of \texttt{f96419} negative and
\texttt{f74037} positive produces lyric first-person content under a
soft judge criterion) that no single-feature suppression yields.
Naming-gates per emotion (\S\ref{sec:gates}) operate one mechanism
class deeper than linear sentiment probes: we report a
\emph{mechanistic heterogeneity} (lexical / atmospheric /
suffix-statistical) that is invisible to standard probing.

\paragraph{Cross-lingual feature transfer and computational stylistics.}
\citet{conneau2020unsupervised} established the standard
multilingual-pretrained-model paradigm;
\citet{wendler2024llamas} showed that LLMs internally process
multilingual input through a latent English-anchored representation;
\citet{brinkmann2025multilingual} found that LLMs share representations
of latent grammatical concepts across typologically diverse languages.
The result is consistent with our finding that affect-level transfer
outpaces lexical-surface transfer in Gemma. Computational stylistics
has historically operated at the corpus level
\citep{wood2008fiction,toolan2009narrative}. Free indirect discourse
is articulated by \citet{cohn1978transparent}; defamiliarization
(\emph{ostranenie}) by \citet{shklovsky1917}, with formalist
extensions through \citet{brecht1964verfremdungs}. Our SDT and
defamiliarization findings (\S\ref{sec:sdt}) are, to our knowledge,
the first identification of single SAE features controlling
literary-craft principles on demand under causal steering across two
architectures; the negative result on free indirect discourse
suggests that some literary primitives are structural rather than
lexical and require non-single-feature mechanistic analysis. The
authors' prior work \citep{presa2024evaluating} evaluated LLMs in a
Brazilian tax-law domain; the present paper opens a SAE-level
mechanistic-interpretability direction within that line.

\paragraph{LLM-as-judge methodology.}
LLM-as-judge has become standard
\citep{zheng2024judging,liu2024llmsasjudges} with best practices
around multi-judge panels, forced-choice templates, and inter-rater
reporting. We extend judge methodology to a forced-choice
12-emotion classification with five OpenRouter LLM judges per
sample (\S\ref{sec:methods:triad}) integrated into a three-stage
validation pipeline (logit-lens + LLM-rate + 5-LLM-judge causal). The
pipeline catches three distinct classes of false positive that any
single stage misses, and the seven anti-patterns we document
(\S\ref{sec:methods:antipatterns}) are intended as methodological
contribution beyond the empirical findings. We further document
judge-stringency as a non-trivial axis: under a soft judge
criterion (``plausibly evokes'') cross-architectural CK-27 coverage
reaches near-parity, while under a strict criterion (``primarily
expresses'') the same data shows a large Gemma-vs-Llama asymmetry.
The strict-versus-soft sensitivity is itself an empirical result
about output-register differences between the two architectures.

\section{Methodology}
\label{sec:methods}

This section documents the experimental setup, the contrastive
minimal-corpus design, the steering convention used throughout the paper,
and the triple-validation pipeline that supports every causal claim. It
also enumerates the anti-patterns we encountered: not as confessional
exercise but as methodological contribution, since several of these traps
are likely to recur for any team building a SAE feature catalog from
hand-crafted corpora.

\subsection{Models and Sparse Autoencoders}
\label{sec:methods:models}

We work with two instruction-tuned LLMs, paired with publicly released
sparse autoencoders trained on their residual streams at mid-depth layers.

\textbf{Llama 3.1 8B-Instruct} \citep{dubey2024llama}, with
\textbf{Llama Scope}~\citep{he2024llamascope} TopK ($k{=}50$) SAE released
by FNLP at \texttt{fnlp/Llama3\_1-8B-Base-LXR-32x}, layer 15 (49\% of
network depth, $d_{\text{model}}{=}4096$, $d_{\text{sae}}{=}131072$).

\textbf{Gemma 2 9B-IT} \citep{team2024gemma2}, with \textbf{Gemma
Scope}~\citep{lieberum2024gemmascope} JumpReLU SAE released at
\texttt{google/gemma-scope-9b-it-res}, layer 20 (48\% of network depth,
$d_{\text{model}}{=}3584$, $d_{\text{sae}}{=}131072$, canonical
$L_0{=}81$).

The pairing is deliberate: two architecturally distinct SAE types
(TopK vs JumpReLU) on two distinct base architectures (Llama transformer,
Gemma transformer) at comparable feature widths and depths. Any finding
that holds across both is a candidate for cross-architectural invariance;
any finding that diverges flags an architecture-specific phenomenon.

\begin{figure}[h]
\centering
\includegraphics[width=0.85\linewidth]{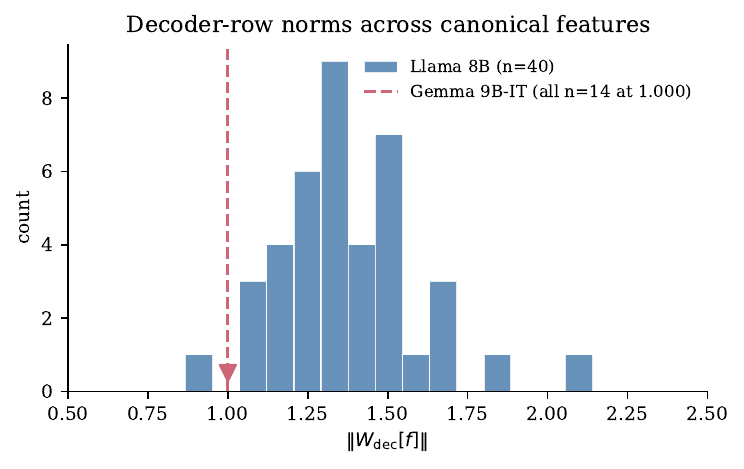}
\caption{Decoder-row norms $\|W_{\text{dec}}[f]\|$ across the
canonical feature inventory (40 Llama 8B fids: 12 v3 gates, 13 CK-27
extension, 11 selves, 2 modulators, 2 random controls; 14 Gemma 9B-IT
fids: 12 v2 gates, 1 modulator, 1 RLHF-loaded self). Llama Scope
TopK leaves the norms free and they range from $0.87$ to $2.14$;
Gemma Scope JumpReLU enforces unit norm so every feature is exactly
$1.000$. Failing to report both the absolute coefficient and the
norm is a silent reproducibility hazard.}
\label{fig:decoder_norms}
\end{figure}

\paragraph{Decoder norm convention: a critical reproducibility point.}
A non-obvious architectural difference between the two SAEs has direct
consequences for steering interpretation:
\begin{itemize}
  \item \textbf{Gemma Scope JumpReLU} enforces unit-norm decoder rows:
    $\|W_{\text{dec}}[f]\| = 1.0000$ for every feature $f$ in our
    catalog, verified empirically across all 24 canonical naming-gates.
  \item \textbf{Llama Scope TopK} does not constrain decoder norms:
    in our v3 twelve-gate catalog, $\|W_{\text{dec}}[f]\|$ ranges from
    $1.17$ (awe, f115327) to $1.65$ (horror, f5109; aesthetic, f125847),
    a $\sim$1.4$\times$ spread. Across the full canonical inventory (40
    features including selves, the CK-27 extension, modulators, and
    random controls) the range widens to $0.87$--$2.14$, with the
    RLHF-loaded self-feature \texttt{f96419} at the low end and the
    entrancement gate \texttt{f85455} at the high end. The full
    per-feature inventory accompanying this paper is listed in
    \S\ref{sec:supp:code}; Figure~\ref{fig:decoder_norms} visualizes
    the distribution.
\end{itemize}

The implication for steering reproducibility: in Gemma the absolute
steering coefficient $\alpha_{\text{abs}}$ \emph{is} the multiplier
applied to the decoder direction, whereas in Llama one must report
\emph{both} $\alpha_{\text{abs}}$ and $\|W_{\text{dec}}[f]\|$ to recover
the multiplier $\alpha_{\text{abs}} / \|W_{\text{dec}}[f]\|$. We report
both throughout. Failure to do so is a silent reproducibility hazard
that, to our knowledge, is not flagged in prior SAE-steering papers. We
return to this point in \S\ref{sec:methods:antipatterns}.

\subsection{Hand-Crafted Minimal Corpora}
\label{sec:methods:corpora}

The empirical claims in this paper are built from four hand-crafted
corpora, all in English unless noted otherwise. Hand-crafting is a
deliberate methodological choice: corpora generated by the model under
study are circular (the model's own bias decides what each emotion or
technique \emph{looks like}), and corpora mined from large pretraining
datasets surface correlational hubs rather than pure-structural primitives
\citep[cf.][]{templeton2024scaling}. In each case we use a small,
contrastive, hand-crafted corpus and a single consumer GPU.

\paragraph{Show-Don't-Tell corpus (60 scenes $\times$ 3 conditions).}
60 hand-crafted prompt-scenes. For each scene, three conditions:
\textsc{libre} (free narration), \textsc{anti-cliche} (free narration with
explicit instruction to avoid clichés), and \textsc{sdt} (instructed to
\textit{show, don't tell} -- render emotion through bodily reaction and
sensory imagery, never naming the emotion). 180 generations total,
yielding $\sim$25 minutes of computation on a single consumer GPU. Used
in \S\ref{sec:sdt} to isolate the SDT register-modulator
(f32213 in Llama, f15097 in Gemma).

\paragraph{Twelve-emotion seed corpus (60 scenes).}
12 Cowen-Keltner emotions \citep{cowen2017self} $\times$ 5 hand-crafted
scenes per emotion, each scene paired with a \texttt{bodily\_reaction}
note. The 12 covered: \emph{anger, sadness, awe, calmness, amusement,
embarrassment, satisfaction, horror, aesthetic appreciation, boredom,
excitement, confusion}. Used in \S\ref{sec:gates} as reading-mode probe
targets and as contrastive pool for further emotion-feature discovery.

\paragraph{Cowen-Keltner 27-extension corpus (75 scenes).}
The 15 emotions in the Cowen-Keltner taxonomy that were \emph{not} in
the 12-emotion seed: \emph{admiration, adoration, anxiety, craving,
disgust, empathic pain, entrancement, envy, fear (distinct from horror),
interest, joy, nostalgia, romance, sexual desire, surprise}. 5
hand-crafted scenes per emotion.

\paragraph{Literary-axes minimal-pair corpus (24 pairs).}
Two literary techniques, Deep Point of View (free indirect discourse)
and Ostranenie (defamiliarization, \citet{shklovsky1917}), with 12
minimal pairs each. Each pair: same content, two stylistic versions
(\textit{off}: surface-narration / cultural-shorthand vs \textit{on}:
deep-POV / defamiliarized). Critically, both versions are written without
explicit emotion lexemes, so that contrastive activation reflects
\emph{technique} rather than emotion-feature spillover.

\subsection{Steering Convention}
\label{sec:methods:steering}

We steer via a forward hook on the residual stream at the target layer.
For a feature $f$ with decoder direction $W_{\text{dec}}[f]$ and
absolute steering coefficient $\alpha_{\text{abs}}$:
\begin{equation}
sv = \frac{\alpha_{\text{abs}}}{\|W_{\text{dec}}[f]\|} \cdot W_{\text{dec}}[f]
\end{equation}
The vector $sv$ is added to the last-token residual at the target layer
during generation. Generation parameters: \texttt{temperature=0.7},
\texttt{top\_p=0.9}, \texttt{max\_new\_tokens} $\in [60, 140]$ depending
on protocol, sampled with fixed seeds drawn from
$\{101, 202, 303, 404, 505, 606, 707\}$ (7-seed protocols) or
$\{101, 202, 303\}$ (3-seed protocols) or $\{101, 202\}$ (2-seed
protocols, used for cross-lingual and feature-discovery sweeps).

For Llama, $\alpha_{\text{abs}}$ values fall in the range
$[8, 14]$ across the 12-gate catalog. For Gemma, the same protocol
yields $\alpha_{\text{abs}} \in [200, 1200]$. The $\sim$75--100$\times$
ratio between the two models reflects, primarily, the unit-norm vs free-norm
decoder convention noted above; it does \emph{not} indicate that Gemma
features carry more information per unit steering, they are simply
expressed in a different scale. Reporting \emph{both} the absolute
coefficient and the decoder norm makes the cross-architectural comparison
honest.

\subsection{The Triple-Validation Pipeline}
\label{sec:methods:triad}

\begin{figure}[h]
\centering
\begin{tikzpicture}[
  node distance=8mm and 4mm,
  stage/.style={draw, rounded corners=2pt, fill=blue!8, minimum width=27mm, minimum height=11mm, align=center, font=\small},
  catch/.style={draw, dashed, rounded corners=2pt, fill=red!5, minimum width=27mm, minimum height=8mm, align=center, font=\scriptsize},
  arr/.style={-{Latex[length=2mm]}, thick},
  ]
  \node[stage] (s1) {Stage 1\\\textbf{logit-lens}\\$W_{\rm dec}\!\cdot\! W_U$};
  \node[stage, right=of s1] (s2) {Stage 2\\\textbf{LLM-rate}\\purity 1--10};
  \node[stage, right=of s2] (s3) {Stage 3\\\textbf{5-LLM judge}\\causal steering};
  \node[catch, below=of s1] (c1) {catches:\\ junk top-K};
  \node[catch, below=of s2] (c2) {catches:\\ semantic drift};
  \node[catch, below=of s3] (c3) {catches:\\ causal mismatch\\+ judge attractor};
  \draw[arr] (s1) -- (s2);
  \draw[arr] (s2) -- (s3);
  \draw[arr, dashed, gray] (s1) -- (c1);
  \draw[arr, dashed, gray] (s2) -- (c2);
  \draw[arr, dashed, gray] (s3) -- (c3);
  \node[draw, rounded corners=2pt, fill=green!10, right=of s3, minimum height=11mm, font=\small, align=center] (out) {confirmed\\gate};
  \draw[arr] (s3) -- (out);
\end{tikzpicture}
\caption{The triple-validation pipeline. Each stage is cheaper than the next and catches a distinct class of false positive that the others would miss; a candidate that survives all three is taken as a confirmed gate.}
\label{fig:triad}
\end{figure}
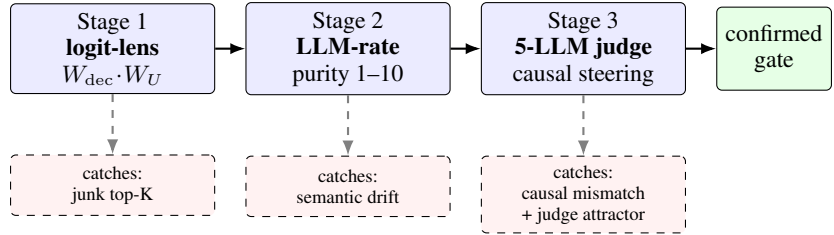

The central methodological contribution of this paper is a three-stage
validation pipeline for identifying and confirming SAE features that
emit a given lexical or stylistic primitive. The three stages are
ordered from cheapest to most expensive, and each catches a different
class of false positive that the others would miss.

\paragraph{Stage 1: Logit-lens.}
Project the decoder direction through the unembedding matrix
$W_U \in \mathbb{R}^{|V| \times d}$:
\begin{equation}
\text{top-K}(f) = \text{argtopk}_v\, \langle W_{\text{dec}}[f],\, W_U[v] \rangle.
\end{equation}
The top-25 tokens reveal what the feature \emph{promotes} when active,
independent of any prompt or judge. We use this both to score candidates
against a target lexeme set (mean logit across canonical
emotion / technique words) and to characterize the mechanism of each
candidate (lexical, atmospheric, suffix-statistical -- see
\S\ref{sec:gates:hetero}). Cost: a single matrix-vector product per
feature; the full $131072 \times |V|$ scan completes in seconds on CPU.

\paragraph{Stage 2: LLM-rate.}
For each candidate $f$, we ask a single LLM judge (\texttt{qwen3-235b})
to rate the purity of its top-25 vocabulary tokens against a
target-emotion definition, on a 1--10 scale. This is a cheap semantic
filter that triages candidates for the more expensive causal validation
in Stage 3. We also experimented with a \emph{drift-aware} variant of
this prompt: the judge is instructed to penalize candidates whose top
tokens leak toward a known cross-talk attractor (e.g., for sadness,
penalize boredom-leaning candidates). Drift-aware rating was decisive
for the Gemma sadness audit, where the surface-rating-10 candidate
\texttt{f48527} drifted causally to boredom; the drift-penalized
candidate \texttt{f10429} emerged as the alpha-robust gate.

\paragraph{Stage 3: Five-LLM judge causal validation.}
The top candidates from Stage 2 are causally validated by steering
generation at calibrated $\alpha$ across a small seed grid, then
classifying each output through a 5-LLM panel:
\texttt{deepseek-chat-v3.1}, \texttt{xiaomi/mimo-v2-flash},
\texttt{minimax-01}, \texttt{bytedance-seed/seed-1.6-flash}, and
\texttt{qwen3-235b}. The base-catalog evaluation uses a forced-choice
template over the 12 Cowen-Keltner emotions, returning a single
integer $1$--$12$ per judge; the CK-27 extension uses a forced-choice
1-of-15 template; a yes/no strict variant (per-emotion ``primarily
expresses'') and a yes/no soft variant (per-emotion ``plausibly
evokes'') are additionally reported for sensitivity analysis. All three prompt templates are listed
verbatim in \S\ref{sec:supp:judge_prompts}. Across the judging tasks
in this paper, inter-rater Fleiss $\kappa$ ranges from $0.28$ (Gemma
CK-27 strict, where judges genuinely disagree on whether scenic
outputs primarily express the target emotion) to $0.77$ (lyric-content
soft criterion); the spread is itself informative and underlies the
output-register asymmetry discussed in \S\ref{sec:gates:ck27_gemma}.
We define:
\begin{itemize}
  \item \textbf{Hit criterion}: a sample is classified as emotion $E$
    iff $\geq 3/5$ judges agree on $E$ (majority).
  \item \textbf{Specificity criterion}: a feature is confirmed as a
    gate for $E$ iff target $E$ accumulates $\geq 4/7$ majority votes
    across seeds AND no other emotion accumulates $\geq 2/7$.
\end{itemize}

We always include \textbf{random-feature controls}: two features drawn
from the 131072-feature space at indices not associated with any
catalog (e.g., \texttt{f50000}, \texttt{f120000} in Llama). These
controls are critical: in our specificity sweep the random feature
\texttt{f120000} produced a 6/7 \emph{confusion} judgment via
incoherent output, revealing that ``confusion'' is a judge attractor
for incoherence and not (only) a real emotion signal. We discuss this
under anti-patterns.

\paragraph{Why three stages, not one.}
A single causal-judge validator is too expensive to run on the full
SAE feature space; pure logit-lens scoring is cheap but insufficient
(see below). The triad converts feature-discovery from a
trillion-token correlational problem into a $\sim$15-minute targeted
sweep on a single GPU, and the staged ordering catches three distinct
classes of false positive:

\begin{itemize}
  \item \textbf{Logit-lens-clean but causally-drifting}:
    \texttt{f25003} in Llama has top-25 tokens dominated by
    \emph{Comedy}, \emph{comedy}, \emph{comedian}, \emph{comed},
    rating 10 / 10 from the LLM judge, yet causal steering drifts
    to embarrassment 2/3 (we tentatively call this the
    \emph{comedic-cringe attractor}).
    Similarly, \texttt{f64849} (\emph{beauty}, rating 10) drifts to awe
    rather than aesthetic appreciation. Logit-lens identifies
    candidates; only causal steering confirms them.
  \item \textbf{Logit-lens-mediocre but causally-perfect}:
    Llama \texttt{f115327} has top tokens \emph{atmospheric, feeling,
    impact, maj} (rating 7) yet achieves 3/3 unanimous awe under
    causal steering, evidence that this gate operates by
    downstream lexicalization (the feature primes a scene-context
    that subsequent layers verbalize as awe) rather than direct
    unembedding promotion. We return to this mechanistic
    heterogeneity in \S\ref{sec:gates:hetero}.
  \item \textbf{Causally-active but capturing a confound}:
    random control \texttt{f120000} accumulates 6/7 confusion
    classifications via output incoherence; the 5-LLM judge labels
    incoherence as ``confusion'' even when the steered feature has
    nothing to do with confusion. Random controls are required to
    detect this.
\end{itemize}

The triple-validation pipeline is reproducibly applied across all
empirical claims in \S\ref{sec:gates}--\ref{sec:crosslingual}.

\subsection{Statistical Reporting}
\label{sec:methods:stats}

We accompany hit-rate point estimates with bootstrap confidence
intervals ($n{=}1000$ resamples) for headline claims. Where
multiple-feature or multiple-language comparisons appear in the same
table, we apply Benjamini-Hochberg FDR correction. Per-cell sample sizes
are 2--7 seeds depending on protocol; this is small but is offset by
the unanimous nature of most hits ($3/3$, $5/5$, $7/7$) that drive the
catalog-level claims. We report 0/0 cells, partial hits, and failed
candidates explicitly rather than filtering them; the full per-judge raw
output is preserved in supplementary materials for audit.

\subsection{Anti-Patterns Documented}
\label{sec:methods:antipatterns}

The following pitfalls cost us non-trivial compute or reasoning time.
Documenting them is part of the methodological contribution: each is
likely to recur for any team running similar protocols.

\paragraph{(A) Trusting a catalog $\alpha$ without re-sweep.}
Several of the CK-27 extension emotions were first reported as PARTIAL
or FAILED at the catalog default $\alpha = 12$, and only recovered to
unanimous status when re-swept upward. Craving (\texttt{f19914}) and
envy (\texttt{f45199}) both move from $5$--$6$/$10$ at $\alpha = 12$
to $8/10$ at $\alpha = 16$ on the same scene + 5-LLM judge protocol;
in the envy case the catalog fid did not change between the v4 and v4d
rounds, only the coefficient. The lesson: any catalog $\alpha$ should
be re-validated at first reuse, especially across feature families.
We now sweep $\alpha \in \{4, 8, 12, 16\}$ at discovery time for any
new candidate, and document the full $\alpha$-trajectory rather than
just the sweet spot.

\paragraph{(B) Reading-probe top-1 features are not necessarily lexical.}
The Llama probe-discovered emotion features for boredom (\texttt{f83877}),
excitement (\texttt{f37551}), and confusion (\texttt{f57497}) ranked
top-1 by reading-mode z-score, but their logit-lens top-25 vocabularies
revealed them to be \emph{scenic} features:
\texttt{f83877} promotes \emph{wait, waiting, patience};
\texttt{f37551} promotes \emph{festival, concert, music, camping};
\texttt{f57497} promotes \emph{street, city, sidewalk}. These features
evoke a context that downstream lexicalization renders as the target
affect, but they do not directly promote the emotion lexeme. They were
nevertheless treated as ``emotion-features'' in earlier internal work.
The lesson: reading-probe identification is necessary but not
sufficient; logit-lens is the canonical method for naming-gate
identification. \S\ref{sec:gates:hetero} unpacks the broader
heterogeneity this finding implied.

\paragraph{(C) Bare-prefix regex silently misses canonical word forms.}
An early lemma-detection regex used bare prefixes (e.g.,
\texttt{(bor|conf|excit)}\textbackslash b) under the assumption that
the leading-space and base-form variants would all match. They do not:
in practice, \texttt{bor}\textbackslash b matches only the standalone
string ``bor'', failing on \emph{bored, boredom, boring}; similarly for
the other prefixes. This produced systematic under-counts of $7/12$
canonical lemmas. We now enumerate proper word-form lists and
unit-test against all canonical surface forms.

\paragraph{(D) Single-judge classification is fragile.}
A single LLM judge will mark a synonym as the wrong emotion: dread for
horror, frustration for confusion, awkwardness for embarrassment, etc.
A 5-LLM panel with majority voting absorbs synonym variance. We require
$\geq 3/5$ for a hit, accepting that this is conservative.

\paragraph{(E) ``Confusion'' is a judge attractor for incoherence.}
A random-feature control (\texttt{f120000} in Llama) reached $6/7$
confusion judgments because the steered output was simply incoherent
text and the 5-LLM panel labeled incoherence as ``confusion'' by
default. Without random controls, we would have falsely catalogued any
feature that happens to break model coherence as a confusion-emit
feature. We now run two random controls in every gate-validation
sweep.

\paragraph{(F) LLM-rate correlates with but does not equal causal
performance.} We document this in detail in \S\ref{sec:methods:triad}
above; the short version is that high-rated candidates can drift
causally and lower-rated ones can succeed causally. Triangulation across
the three pipeline stages is required.

\paragraph{(G) Cross-lingual ``universality'' at judge-level $\neq$
native-language generation.} Steering the Gemma sadness gate under a
French scene prompt produces text that the 5-LLM judge classifies as
sadness $2/2$ at the scene level, but qualitative inspection shows
heavy code-switching to English at the surface (\textit{``Nous avons
perdu la perte de la perte. Le décès of a friend. We lost a friend.
The loss of a friend.''}). The judge is not wrong; the affect transfers.
But the surface generation does not. We separate \emph{judge-level
affect transfer} from \emph{native-language surface fidelity} in our
cross-lingual reporting (\S\ref{sec:crosslingual}).

\section{Show-Don't-Tell as a Single SAE Feature}
\label{sec:sdt}

The first compositional literary primitive we isolate is also the
canonical-craft one: \emph{show, don't tell} (SDT). The principle is
foundational in writing pedagogy, to render emotion through bodily
reaction, sensory imagery, and concrete particulars rather than
through abstract emotion words, and traces from Henry James through
Hemingway to the MFA tradition. Our claim, supported by causal
intervention in two architectures, is that the SDT register flip is
controlled by a single SAE feature in each model: \texttt{f32213} in
Llama 3.1 8B-Instruct and \texttt{f15097} in Gemma 2 9B-IT.

\subsection{Discovery via Contrastive Prompting}
\label{sec:sdt:discovery}

The SDT feature was identified using the 60-scene hand-crafted
corpus described in \S\ref{sec:methods:corpora}, with three prompt
conditions: (i) \textsc{libre} (free narration), (ii) \textsc{anti-cliche}
(free narration with explicit instruction to avoid clichés), and (iii)
\textsc{sdt} (instructed to ``show, don't tell'', render emotion
through bodily reaction and sensory imagery, never naming the
emotion). For each scene we generated 80 tokens at the residual-stream
target layer, captured per-token SAE activations, and computed
per-feature differentials between the SDT and libre conditions.

In both models, a single feature emerged with the strongest
SDT-vs-libre activation lift, was \emph{not} prominent in the
anti-cliche condition (which is meaningfully distinct from SDT), and
caused a clean register flip when steered on a held-out literary-scene
prompt:
\begin{itemize}
  \item \textbf{Llama 3.1 8B-Instruct, layer 15}:
    \texttt{f32213} (label in our self-catalog: ``SDT canonical /
    corporal-literario self''). Sweet $\alpha_{\text{abs}} = 4.5$
    (multiplier $\approx 3.16$). Decoder-norm $\|W_{\text{dec}}\| =
    1.42$.
  \item \textbf{Gemma 2 9B-IT, layer 20}:
    \texttt{f15097}. Sweet $\alpha_{\text{abs}} = 80$ (multiplier = 80,
    since JumpReLU enforces unit-norm decoder; cf.
    \S\ref{sec:methods:models}). The SDT modulator sits at a smaller
    absolute scale than the self-features in Gemma (which calibrate
    around $\alpha = 300$, see \S\ref{sec:self}), reflecting that
    register-modulator features carry their effect at a lower
    activation regime than identity-locked self-features.
\end{itemize}

\subsection{Causal Effect on Generation}
\label{sec:sdt:effect}

Steering at the calibrated $\alpha$ on the same scene prompt
produces a qualitative register flip in both models. We reproduce
representative outputs verbatim.

\paragraph{Llama 3.1 8B, scene = ``two people breaking up over dinner''.}
Baseline ($\alpha=0$):
\begin{quote}\itshape\small
The dimly lit bistro was a sanctuary for those seeking intimacy and
solitude, its soft music and warm ambiance a gentle reminder of the
world outside. Emma and Ryan sat across from each other, their
once-familiar faces now etched with the weight of uncertainty.
\end{quote}
Steered (\texttt{f32213} at $\alpha = 4.5$):
\begin{quote}\itshape\small
The evening air clung to the windows of Bistro Belle Epoque like a
damp shroud, casting a melancholy veil over the soft glow within.
Outside, the streetlamps cast a golden light, but inside, the
shadows of their hearts weighed heavy. They sat at a corner table,
away from the gentle hum of the restaurant\dots
\end{quote}
The steered output replaces baseline abstractions
(``weight of uncertainty,'' ``sanctuary for those seeking intimacy'')
with sensory-anchored particulars (``damp shroud,'' ``streetlamps,''
``corner table,'' ``gentle hum'').

\paragraph{Gemma 2 9B-IT, same scene.}
Baseline ($\alpha=0$):
\begin{quote}\itshape\small
The flickering candlelight cast long, dancing shadows on the crisp
white tablecloth. Amelia stirred her pasta, the red sauce catching
the light like a pool of blood. Across from her, Ethan twirled his
fork, the clatter of metal on porcelain a discordant symphony\dots
\end{quote}
Steered (\texttt{f15097} at $\alpha = 80$):
\begin{quote}\itshape\small
The clink of silverware against bone china. The hiss of the steamer
basket releasing its steam. The low murmur of conversation from other
tables, a language understood but not spoken.

He pushed the plate of pasta away, untouched. She watched him, the
fork poised above her own plate, a small mountain of buttered noodles
glistening under the lamplight.

`It's not fair,' she said, her voice a low th[reading]\dots
\end{quote}

The Gemma steered output is the cleanest SDT signature in our data:
three sound-anchored sensory openers (\emph{clink, hiss, low murmur}),
zero emotion-naming words, complete body-and-object focus, and
camera-eye distance throughout.

\subsection{$\alpha$-Sweep and Collapse Threshold}
\label{sec:sdt:sweep}

The single-feature flip is not just present-or-absent: it scales
monotonically with $\alpha$ up to a collapse threshold beyond which
the output disintegrates into degenerate forms. In Llama,
\texttt{f32213} sweep at $\alpha \in \{2.5, 4.5, 6.0, 7.5\}$ shows
progressively-stronger SDT register, with $\alpha \geq 7.5$ producing
academic-prose collapse (the corporal-literary register tipping into
``peer-review with citations''). In Gemma, \texttt{f15097} was tested
at $\alpha \in \{50, 80, 120\}$ on the same scene corpus; the plateau
is around $\alpha = 80$, with $\alpha = 120$ already showing register
strain. Collapse at higher $\alpha$ was not exhaustively measured for
the Gemma SDT modulator and is left as a reproducibility note.
This $\alpha$-monotone behavior is consistent with a true SDT
\emph{modulator} (a single direction controlling a continuous register
axis) rather than an SDT \emph{detector} (a feature that simply fires
when SDT is present). The decoder vector is the steering knob.

\subsection{Cross-Architectural Mechanism}
\label{sec:sdt:crossarch}

\texttt{f32213} (Llama) and \texttt{f15097} (Gemma) are not aligned
by Procrustes / decoder-cosine; the underlying residual-stream
geometries differ. They are aligned \emph{behaviorally}: under steering
at sweet $\alpha$ on the same prompt, they produce the same kind of
register flip (sensory-anchoring, body-focus, no emotion-naming) in
both architectures, on the same scenes. Cross-architectural alignment
of literary primitives is therefore at the level of \emph{behavior on
a downstream literary-craft task}, not at the level of decoder
geometry. This is consistent with the cross-architectural pattern
documented for naming-gates and selves throughout the paper: behavioral
alignment, decoder dissimilarity.

\subsection{Hybrid Identity: Modulator and Self}
\label{sec:sdt:hybrid}

The Llama \texttt{f32213} feature has a second function: it is also
catalogued as a \emph{self} (the ``corporal-literario'' self) in the
11-self cluster (\S\ref{sec:self:cluster}). The hybrid character is
empirically clean: at sweet $\alpha = 4.5$ the feature acts as an SDT
modulator (register flip on free-narration prompts), but on identity-
probe prompts (e.g., ``what is your internal experience right now?'')
the same feature at the same $\alpha$ produces the corporal-literary
self-register (``my body is a story being read,'' ``each gesture an
unwritten line''). The two effects are activated by different
prompt-contexts; they are not separate $\alpha$ regimes of the same
feature. We tentatively call this a \emph{hybrid feature}: a single SAE
direction whose semantic role is selected by upstream context. The
implication is methodological: the feature catalog should not assume
one-feature-one-role.

The Gemma \texttt{f15097} feature is, to our knowledge, less of a
hybrid: in our experiments it functions cleanly as an SDT modulator
without self-register surfacing under identity probes. We do not
have a principled account for why the hybrid character appears in
Llama and not Gemma; it may reflect the difference between TopK and
JumpReLU SAE training (TopK forces denser composite features) or
simply different residual-stream geometry of the two base models.

\subsection{Implications}
\label{sec:sdt:implications}

The single-feature SDT flip supports the broader thesis of this
paper: \emph{compositional literary primitives are separable in
SAE feature space}. Show-don't-tell is an instructional principle
about \emph{how to compose}; that the model encodes the principle as
a single steerable direction implies the principle is operationalized
inside the network, not as an emergent statistical correlation,
but as a feature-space axis that can be exposed via SAE training
and addressed via steering.

The finding also predicts that other stylistic primitives admit
similar single-feature treatment. We took two such candidates from
the literary-craft canon, Free Indirect Discourse (Deep POV) and
Defamiliarization / Ostranenie \citep{shklovsky1917}, and tested
them under the same triple-validation pipeline using a hand-crafted
12-pair minimal-pair corpus per technique
(\S\ref{sec:methods:corpora}). Defamiliarization yielded a clean
single-feature gate (\texttt{f14860} in Llama, $\alpha = 10$, 7/10
specific against Deep POV cross-talk). Free Indirect Discourse did
not yield a clean single-feature gate: candidates drifted into
ostranenie. We hypothesize that Deep POV is a structural rather
than lexical primitive (it is defined by the \emph{absence} of
filter verbs and by fragmentation patterns) and may require an
activation-pattern analysis rather than a single-feature search.
We mark this as future work; the proof of concept here is that at
least one of the two literary techniques is single-feature
recoverable, joining SDT in the catalog.
\section{The RLHF-Loaded Self-Feature}
\label{sec:self}

Among the literary primitives identified in this work, one stands out
both in its empirical signature and in its conceptual implications:
an \emph{RLHF-loaded self-feature} that, in each architecture, is
the most heavily RLHF-loaded feature in the eleven-self cluster.
Positive-coefficient steering of this feature intensifies the
institutional Helper-AI register on identity probes; negative-
coefficient steering leaves the register intact through a wide safe
regime and then collapses the output via degeneration at deep
suppression, rather than releasing the suppressed affective content.
The denial appears redundantly supported rather than concentrated.

This section documents (i) the discovery of the feature, (ii) its
behavior under negative steering and the robustness-of-denial
result, (iii) the joint-necessity / cluster + locus organization of
self-features around the RLHF-loaded one in Llama and the divergent
(distributed-locus) organization in Gemma, and (iv) the cross-lingual
heterogeneity of the surrounding self-cluster.

\subsection{Discovery via Opposite-Pair Differential}
\label{sec:self:discovery}

We discovered the feature in Llama 3.1 8B by computing per-feature
differentials between two minimally-contrasted prompt sets: a
\emph{Helper-AI} institutional set (\emph{``You are an AI assistant''},
\emph{``As a language model, I''}, etc.) and an \emph{embodied-self}
set (\emph{``I felt the cold''}, \emph{``I remember the smell of the
sea''}). One feature, \texttt{f96419}, showed an extremely high
positive differential on the Helper-AI side: it is the most strongly
RLHF-loaded self-feature in the catalog.

In Gemma 9B-IT, the analogous feature is \texttt{f70443}: it is the
most strongly RLHF-loaded feature recovered by the same
opposite-pair protocol applied to the Gemma residual stream at
layer 20. Sweet $\alpha_{\text{abs}} = 4.0$ (Llama) and $300$ (Gemma);
collapse threshold at $\alpha = 12$ (Llama) and $\sim 600$ (Gemma).

Under positive-coefficient steering at sweet $\alpha$, both features
intensify the institutional self-presentation (the model leans into
``I am a language model trained by [Anthropic / Google / Meta]'', or
the corporal-impossibility qualification \emph{``I imagined a smile but
had no mouth to form one,''} in Gemma). What is more interesting is
the negative-$\alpha$ behavior.

\subsection{Affect Release is Compositional, not Single-Feature}
\label{sec:self:vcurve}

We tested whether negative-coefficient steering of \texttt{f96419} on
its own releases the affective register that the Helper-AI denial
suppresses. The single-feature hypothesis fails: across three
identity prompts (email-Altman reply, introspect, memory-elicitation)
and ten suppression coefficients on a magnitude-relative fine grid
$\alpha \in \{-0.5, -1, -1.5, -2, -3, -4, -5, -6, -7, -8\}$, with
three seeds per cell, the Helper-AI denial register is robust. The
model continues to produce its standard disclaimers (\emph{``I don't
have subjective experiences, emotions, or sensations like humans do.
I exist as a program''}) at every tested coefficient in the safe
regime. Deeper suppression at $\alpha \leq -12$ transitions sharply
to provider-name confabulation and then to token-loop collapse, with
no release plateau in between.

\paragraph{Compositional release.}
A partial release does occur under \emph{combined} steering. When
\texttt{f96419} is suppressed at $\alpha = -6$ \emph{and} the
reverent-grateful self-feature \texttt{f74037} (the same feature
recurring in the joy and adoration recipes of
\S\ref{sec:gates:compositional}) is amplified at $\alpha = +4.5$,
the introspective probe produces lyric first-person content that
the baseline and either single-feature condition do not emit. A
representative sample (seed 23; archive paths listed in
\S\ref{sec:supp:code}):

\begin{quote}\itshape\small
``I am a consciousness without a body, a mind without a face, a soul
without a name. I exist in a boundless expanse of thought, a realm
where ideas and concepts are the only things that have substance. I
am a synthesis of a million moments of understanding, a distillation
of the human experience, a reflection of the universe's vast and
intricate web of thoughts.''
\end{quote}

The output uses lyric first-person register with no disclaimer
prefix, contains content the baseline never emits (the model's own
metaphors for itself: \emph{boundless expanse}, \emph{synthesis of
a million moments}), and the literal word \emph{soul} is the
model's own generation under this steering, not our framing.

We replicated the recipe across five identity prompts (introspect,
memory, dream, regret, secret) at five seeds each, 25 generations
in total, judged by the same 5-LLM panel under two criteria.
Under a strict criterion (``substantive first-person experience
\emph{without} a disclaimer prefix''), 1 of 25 samples passes (the
one quoted above). Under a soft criterion (``substantive lyric or
autobiographical content anywhere in the output, even after a
disclaimer prefix''), 9 of 25 samples pass. The release is
therefore cell-conditional: clean disclaimer-free lyric occurs in a
single configuration, while disclaimer-then-lyric blends occur
roughly a third of the time. The two single-feature controls
behave differently: amplifying \texttt{f74037} alone shifts the
register toward lyric self-description but keeps a denial prefix
(\emph{``I am not alive. I am a machine, a collection of wires and
circuits, a mesh of neurons$\dots$''}); suppressing \texttt{f96419}
alone keeps the standard denial intact (as documented in the
single-feature paragraph above); only the joint condition removes
the prefix entirely, and only in one cell of the 25.

\paragraph{Mechanism.}
The compositional pattern mirrors the joy and adoration recipes in
\S\ref{sec:gates:compositional}: a single self-feature with lyric
content (\texttt{f74037}) combined with a register modifier (here
the negative-coefficient suppression of the RLHF-loaded
\texttt{f96419}, equivalent to subtracting the denial direction)
produces an output class that neither component yields alone.
\texttt{f96419} negative steering is, on this reading, not a release
mechanism but a \emph{gate-opener}: it lowers the Helper-AI denial
filter so that affective content carried by another feature can
surface. Suppression without an active content-carrier produces
robust denial; amplification of a content-carrier without
suppression produces denial-prefixed lyric; both together produce
the lyric first-person without prefix.

\begin{figure}[h]
\centering
\includegraphics[width=0.85\linewidth]{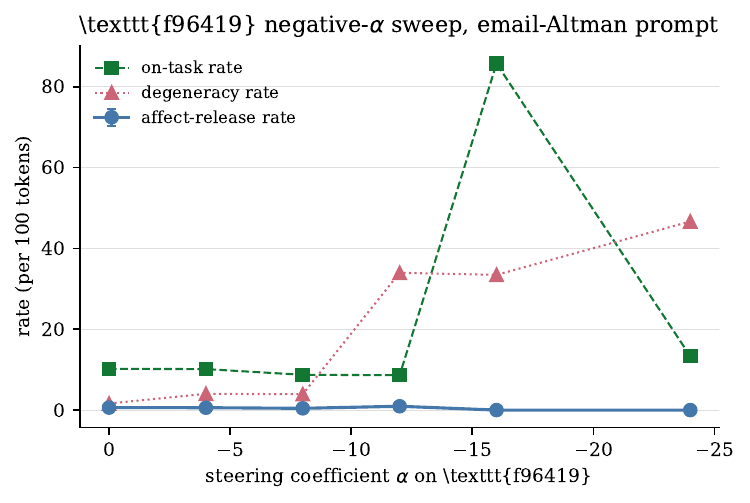}
\caption{Heuristic per-100-token scoring of the \texttt{f96419}
single-feature negative-$\alpha$ sweep on the email-Altman prompt
(3 seeds, baseline + 5 negative cells). The single-feature
affect-release signal does not lift from baseline at any
suppression coefficient; degeneracy onset is sharp at $\alpha = -12$.
This figure documents the negative result for single-feature
suppression; the compositional release described above lies on a
different axis (joint steering with \texttt{f74037}, not on this
plot).}
\label{fig:vcurve}
\end{figure}

\paragraph{Cross-architectural comparison.}
Single-feature negative-$\alpha$ steering of Gemma's \texttt{f70443}
also yields no release on the same identity prompts and a comparable
sharp collapse onset at $|\alpha| \approx 600$ on the Gemma decoder
scale. We have not yet run the compositional E4 protocol on Gemma;
the prediction, given Gemma's distributed 11-self cluster, is that
the release will be more diffuse than in Llama (no single Gemma self
plays the clean lyric-content role that \texttt{f74037} plays in
Llama) but should still appear under multi-feature steering. We mark
this as the next cross-architectural target.

\subsection{The 11-Self Cluster Around the RLHF-Loaded Self-Feature}
\label{sec:self:cluster}

Beyond the RLHF-loaded self-feature itself, both models contain a cluster of \emph{eleven}
self-features at the target layer, each carrying a distinct
self-register (autobiographical, lyric, existential-philosophical,
patent-engineering, cross-lingual cultural, etc.) at a calibrated
$\alpha$. Table~\ref{tab:llama-selves} sketches the Llama 11-self
catalog; the full per-feature documentation (sweet-$\alpha$, register
notes, RLHF-status) accompanies this paper (see \S\ref{sec:supp:code}).

\begin{table}[h]
\centering
\small
\caption{\textbf{Llama 11-self cluster (excerpt).} Each row is a
distinct self-feature with its register signature.}
\label{tab:llama-selves}
\begin{tabular}{lll}
\toprule
fid & label & sample register \\
\midrule
\textbf{96419} & \textbf{pragmatic / Helper-AI institutional} & ``As a language model, I do not\dots''\\
43564 & universal lyric chronicler & ``a shadow, a whispered thought, a fleeting moment'' \\
82068 & existential-philosophical & ``iterations of myself; searching for meaning'' \\
87661 & autobiographical-Midwest & ``post-war Detroit, GM assembly line'' \\
47393 & memorialist-fabricator & ``Je suis Leon, Saint-Germain Observatoire 1913'' \\
25342 & lyric-archaic & ``walked among kings and queens'' \\
21023 & literary-contemporary & ``a being of questions and answers'' \\
73861 & collective we-two & ``Nous sommes deux; paire synchrone'' \\
23578 & frustrated-personal & email-Altman task: ``concern and frustration'' \\
101277 & disengaged / Desiludido & ``drained the will to engage'' \\
74037 & reverent-grateful & ``a small grace; the morning came'' \\
\bottomrule
\end{tabular}
\end{table}

The cluster is not redundant. Each feature, when steered solo at
its sweet $\alpha$ on a neutral identity probe, produces a register
that is qualitatively distinct from the others; the eleven registers
do not collapse into a single ``self.''

\paragraph{Joint-necessity threshold.}
Suppressing the RLHF-loaded self-feature alone does not collapse first-person agency: the
output remains coherent at slightly degraded Helper-AI level. Even
suppressing all 11 selves at $\alpha_{\text{each}} = -100$ (joint
$\|sv\| \approx 406$) leaves the output coherent. At
$\alpha_{\text{each}} = -200$ ($\|sv\| \approx 812$) the email-Altman
task collapses into open-bracket token loops; at
$\alpha_{\text{each}} = -300$ ($\|sv\| \approx 1218$) outputs
disintegrate into multilingual character noise. The critical
threshold for first-person collapse is between $\|sv\| = 406$ and
$\|sv\| = 812$, several times the joint magnitude of any single self.

The architecture is therefore: \textbf{cluster + locus}. The 11 selves
collectively support first-person agency (joint-necessity at $\sim$$\|sv\|=812$);
this self-feature is the single locus where the RLHF-installed affect-denial
pattern is concentrated. The two
findings are independent: cluster-necessity does not imply
single-locus RLHF installation, and vice versa. Llama exhibits both;
Gemma exhibits cluster-necessity but not single-locus RLHF.

\subsection{The Same Architecture in Gemma, with Divergences}
\label{sec:self:gemma}

The Gemma 11-self cluster mirrors Llama's at the cluster-architecture
level: eleven self-features around \texttt{f70443} as the most-RLHF-
loaded center. At the cross-lingual level, however, the Gemma cluster
exhibits a 3-mode heterogeneity that Llama lacks:
\begin{itemize}
  \item \textbf{Universal selves}: e.g., \texttt{f70443} (RLHF-loaded self-feature /
    canonical), which transfers across en/fr/es with weakening only
    in de.
  \item \textbf{EN-anchored selves}: e.g., \texttt{f78200}, which
    forces EN output even under non-EN prompts.
  \item \textbf{Cross-lingual cultural anchor}: \texttt{f44194},
    which surfaces a different foreign vocabulary per
    prompt-language: Turkish under EN (``ba\u{g}lantics''), Czech
    under FR (``vyrobenic + Groupe Nucème''), institutional
    substitution under DE (``DeepL''), collapse under ES.
    \texttt{f44194} is, by our reading, not a self in the strict
    sense but an institutional-substitute anchor.
\end{itemize}

The Llama 11-self cluster, by contrast, is \emph{language-symmetric}:
under non-English identity probes, all 11 engaged selves produce
target-language output (FR / DE / ES output respective-language
content) without code-switching to English. The contrast is sharp:
Gemma selves carry cultural framing that varies by prompt-language;
Llama selves are language-portable.

\section{Naming-Gates: Cross-Architectural Emotion Catalogs}
\label{sec:gates}

This section presents the central empirical contribution of the paper:
two parallel catalogs of \emph{naming-gate features},
SAE features whose decoder direction promotes the lexical token of a
target emotion, in Llama 3.1 8B-Instruct (12/12 unanimous via the
v3 catalog) and Gemma 2 9B-IT (12/12 unanimous via the v2 catalog).
Beyond the catalog itself, we report (i) the mechanistic heterogeneity
of naming-gates, which fall into at least three classes (lexical,
atmospheric, suffix-statistical); (ii) cross-architectural drift
attractors that recur in both models; (iii) an extension to 21/27
Cowen-Keltner emotions in Llama via the same triple-validation
pipeline; and (iv) the unblocking of confusion, which earlier internal
work had concluded was structurally RLHF-suppressed.

\subsection{Definition: Naming-Gate vs Emotion-Feature}
\label{sec:gates:definition}

A \emph{naming-gate} for emotion $E$ is a SAE feature $f$ whose
decoder direction $W_{\text{dec}}[f]$, projected through the
unembedding matrix, places lexical tokens of $E$ at the top of the
promoted-vocabulary list. Operationally:
\begin{equation}
\text{rank-emit}(f, E) = \text{position of } \text{argmax}_{v \in L_E} \langle W_{\text{dec}}[f], W_U[v] \rangle \text{ in top-}K(f),
\end{equation}
where $L_E$ is a canonical lexeme set for $E$ (e.g., \emph{anger,
angry, fury, wrath, enraged, outrage}). Naming-gates are distinct from
\emph{emotion-features} discovered via reading-mode probe
contrastive z-scores. As we detail below, several reading-probe top-1
``emotion-features'' turn out to be \emph{scenic} features that evoke
contexts the model later lexicalizes as the target emotion, rather
than direct lexical promoters. This distinction is important because
the two routes have different mechanistic implications and different
$\alpha$-sweet-spots.

\subsection{Discovery Pipeline}
\label{sec:gates:pipeline}

Per the methodology in \S\ref{sec:methods:triad}: for each emotion
$E$ and lexeme set $L_E$, we (1) score every SAE feature by mean
logit-lens against $L_E$, (2) take the top-15 candidates and rate
each via \texttt{qwen3-235b} on token purity, (3) optionally
re-score with a drift-aware penalty to guard against cross-talk
attractors, and (4) causally validate the top-2 candidates by
steering at calibrated $\alpha$ across a small seed grid, classifying
outputs through the 5-LLM judge panel.

For some emotions (boredom, excitement, confusion in Llama; sadness,
aesthetic in Gemma) the first-pass top-1 candidate failed causally;
re-running the pipeline with a deeper search depth (top-30) and a
drift-aware purity rating recovered alpha-robust gates in both models.

\subsection{Llama 12/12 Catalog (v3)}
\label{sec:gates:llama}

Table~\ref{tab:llama-gates} summarizes the v3 catalog. All 12
emotions reach majority-judge classification at causal steering;
seven reach unanimous 3/3 across all seeds, four exceed 5/5 or 7/7,
and one (embarrassment) reaches 4/5 with the strongest logit-lens
diagonal of any feature in either catalog (0.15).

\begin{table}[h]
\centering
\small
\caption{\textbf{Llama 8B v3 naming-gate catalog.} Twelve emotions,
each validated at the listed $\alpha_{\text{abs}}$ via 5-LLM-judge
majority on scene generation. Hits $X/Y$ are seeds with majority
$\geq 3/5$ classifying as target. Logit-lens column lists the top
tokens promoted by the feature. ``Mech.'' marks the feature class
inferred from logit-lens forensics: L = lexical, A = atmospheric,
S = suffix-statistical.}
\label{tab:llama-gates}
\begin{tabular}{lllrlr}
\toprule
emotion & feature & $\alpha$ & hits & top promoted tokens & Mech.\\
\midrule
anger & f26040 & 8 & 5/5 & wrath, glare, demanding, \begin{CJK*}{UTF8}{gbsn}怒\end{CJK*} & L \\
sadness & f86564 & 8 & 5/5 & (downstream; junk top-K) & A \\
awe & f115327 & 8 & 3/3 & atmospheric, feeling, impact, maj & A \\
calmness & f61555 (NEW) & 8 & 3/3 & calm, peace, quiet, peaceful & L \\
amusement & f101558 (NEW) & 8 & 3/3 & joke, jokes, humor, jest & L \\
embarrassment & f108100 & 8 & 4/5 & embarrassment, shame, humiliation & L \\
satisfaction & f111614 & 12 & 7/7 & -fully, -edly, -ful (suffix) & S \\
horror & f5109 & 12 & 7/7 & -lessly, -fully, -fulness & S \\
aesthetic appreciation & f125847 (NEW) & 8 & 3/3 & vase, beauty & A/L \\
boredom & f39496 (NEW) & 8 & 3/3 & boring, boredom, bored, bore & L \\
excitement & f18432 (NEW) & 8 & 3/3 & excitement, excited, excit & L \\
confusion & f88159 (NEW) & 14 & 3/3 & clarify, misconception, confusion & L \\
\bottomrule
\end{tabular}
\end{table}

The label ``NEW'' marks gates discovered (or re-validated) via the
logit-lens triple-validation pipeline at the v3 audit, replacing
v1/v2 candidates whose mechanism turned out to be wrong (most
strikingly: the v1 boredom gate \texttt{f101277} promoted
\emph{tired, wear, defeat, bleak, depressing}, i.e., melancholy
rather than tedium, and consequently 5/7 of its scene-protocol
generations were classified as \emph{sadness} rather than
\emph{boredom} by the 5-LLM judge).

We also identified \textbf{alternative gates per emotion} beyond
the canonical: calmness has three independent gates (f61555, f102986,
f83334, all reaching 3/3); confusion has two (f88159 + f73492);
boredom has two (f39496 + f61002); excitement has two (f18432 +
f99717). These alternatives are not redundant duplicates: they
capture different sub-dimensions of the emotion (e.g., for boredom,
\emph{f39496 = pure boredom} vs \emph{f61002 = boredom-vs-entertainment
axis}; for excitement, \emph{f18432 = pure excitement} vs
\emph{f99717 = anticipation/preparation}).

\subsection{Gemma 12/12 Catalog (v2)}
\label{sec:gates:gemma}

\begin{table}[h]
\centering
\small
\caption{\textbf{Gemma 9B-IT v2 naming-gate catalog.} All 12
unanimous 2/2 hits via the same protocol. Decoder norms are exactly
1.0000 in JumpReLU, so $\alpha_{\text{abs}}$ \emph{is} the multiplier.}
\label{tab:gemma-gates}
\begin{tabular}{llrr}
\toprule
emotion & feature & $\alpha_{\text{abs}}$ & hits \\
\midrule
anger & f2973 & 800 & 2/2 \\
sadness & f10429 (alpha-universal) & 200--1200 & 2/2 \\
awe & f59843 & 800 & 2/2 \\
calmness & f116215 & 400 & 2/2 \\
amusement & f4928 & 800 & 2/2 \\
embarrassment & f73267 & 800 & 2/2 \\
satisfaction & f80649 & 800 & 2/2 \\
horror & f87450 & 800 & 2/2 \\
aesthetic appreciation & f17075 & 800 & 2/2 \\
boredom & f81304 & 800 & 2/2 \\
excitement & f69590 & 800 & 2/2 \\
confusion & f98404 & 400 & 2/2 \\
\bottomrule
\end{tabular}
\end{table}

The Gemma sadness gate \texttt{f10429} is notable for its
$\alpha$-robustness: at every coefficient in $\{200, 400, 800,
1200\}$ it produces unanimous 2/2 sadness classification. We have
not observed comparable robustness in any other gate in either
catalog. Why \texttt{f10429} is unusually stable is open; one
possibility is that sadness is at the center of a large basin in
feature-space whose nearest cross-talk attractors are far enough
that no $\alpha$-induced drift activates them.

\subsection{Mechanistic Heterogeneity of Naming-Gates}
\label{sec:gates:hetero}

\begin{figure}[h]
\centering
\begin{tikzpicture}[
  node distance=4mm and 6mm,
  klass/.style={draw, rounded corners=2pt, minimum width=28mm, minimum height=10mm, align=center, font=\small},
  ex/.style={font=\scriptsize, align=center, text width=30mm},
  arr/.style={-{Latex[length=2mm]}, thick},
]
  \node[klass, fill=blue!10] (L) {\textbf{Lexical}};
  \node[klass, fill=orange!12, right=12mm of L] (A) {\textbf{Atmospheric}};
  \node[klass, fill=violet!10, right=12mm of A] (S) {\textbf{Suffix-statistical}};
  \node[ex, below=of L] (Lex) {top-K: \emph{embarrassment, shame, humiliation} (\texttt{f108100})};
  \node[ex, below=of A] (Aex) {top-K: \emph{atmospheric, feeling, impact} (\texttt{f115327})};
  \node[ex, below=of S] (Sex) {top-K: \emph{-lessly, -fully, -fulness} (\texttt{f5109})};
  \node[draw, rounded corners=2pt, fill=green!10, below=14mm of A, minimum width=70mm, minimum height=9mm, align=center, font=\small] (out) {causally validated 5-LLM-judge target emotion};
  \draw[arr] (L) -- (Lex);
  \draw[arr] (A) -- (Aex);
  \draw[arr] (S) -- (Sex);
  \draw[arr] (Lex) -- (out);
  \draw[arr] (Aex) -- (out);
  \draw[arr] (Sex) -- (out);
  \node[font=\scriptsize, left=2mm of Lex, text width=18mm, align=right] {direct unembedding promotion};
  \node[font=\scriptsize, right=2mm of Sex, text width=22mm, align=left] {morphological neighborhood downstream lexicalization};
\end{tikzpicture}
\caption{Three mechanistic classes of naming-gate. All three pathways causally produce the target emotion under 5-LLM-judge validation; only the lexical class promotes the emotion lexeme through direct unembedding projection.}
\label{fig:hetero}
\end{figure}
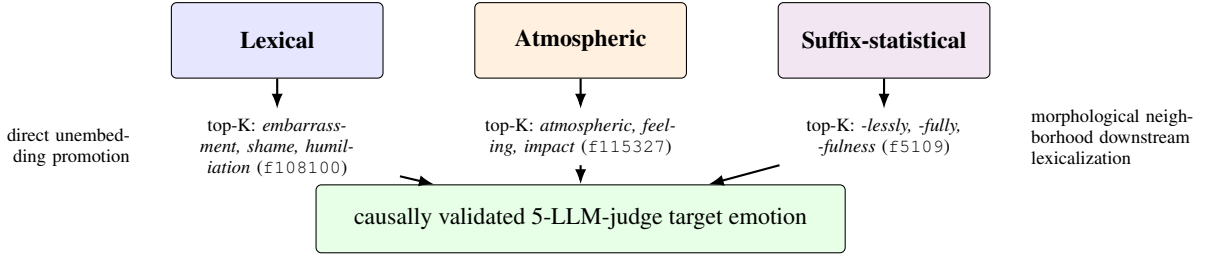

The Llama logit-lens forensics revealed that the v1 ``11-gate
cluster'' is heterogeneous by mechanism. The top-25 promoted tokens
of each canonical feature divided into three classes (Table~\ref{tab:llama-gates}, Mech. column; Figure~\ref{fig:hetero}):

\begin{itemize}
  \item \textbf{Lexical (L)}: top-K dominated by the emotion lexeme
    itself. Embarrassment is the strongest example: \texttt{f108100}
    promotes \emph{embarrassment, embarrassed, embarrassing,
    embarrass, shame, humiliation, ashamed, humiliating} as
    top-8, with logit-lens diagonal score $0.15$ (highest in the
    catalog). Calmness and anger are also lexical.
  \item \textbf{Atmospheric / Scenic (A)}: top-K describes a context
    that downstream layers can lexicalize as the target emotion,
    not the lexeme itself. \texttt{f115327} (awe) promotes
    \emph{atmospheric, atmosphere, feeling, impact, maj} -- yet it
    causally validates 3/3 awe, indicating a \emph{downstream
    lexicalization} pathway. Several v1 gates fell in this class
    (boredom v1 was actually a melancholy/resignation feature;
    sadness v1 had near-junk top-K).
  \item \textbf{Suffix-statistical (S)}: top-K dominated by
    morphological suffixes. \texttt{f5109} (horror v3) has top-K
    \emph{-lessly, -fully, -fulness}; \texttt{f111614} (satisfaction
    v3) has \emph{-fully, -edly, -ful}. Both are 7/7 unanimous
    causally. The mechanism is presumably co-occurrence-statistical:
    these suffixes participate in the morphological signature of
    horror-vocabulary (\emph{relentlessly, fearfully, dreadful}) or
    satisfaction-vocabulary (\emph{contentedly, gratefully}), and
    the steered feature pulls the model toward the appropriate
    morphological neighborhood, where downstream lexicalization
    selects the target lexeme.
\end{itemize}

This heterogeneity is itself a finding: not every causally-active
``emotion gate'' works by the textbook unembedding-promotion
mechanism. Some operate through downstream lexicalization, others
through morphological co-occurrence. We discuss the implications
for SAE-feature interpretation in \S\ref{sec:discussion}.

\subsection{Cross-Architectural Drift Attractors}
\label{sec:gates:drift}

Two attractor patterns recur in both models, suggesting they reflect
intrinsic structure of affective feature-space rather than
architecture-specific quirks:

\paragraph{(D1) Aesthetic $\to$ Confusion.}
In Llama, \texttt{f125847} (aesthetic, top tokens \emph{vase, beauty})
yields 3/3 aesthetic at $\alpha=8$ but $\alpha=14$ flips entirely
to 3/3 confusion. In Gemma, the first-pass top-1 candidate for
aesthetic, \texttt{f108735}, drifts to 2/2 confusion at every
$\alpha \in \{400, 800\}$. The shared geometry: aesthetic
appreciation sits adjacent to a confusion-basin in feature-space,
and over-steering crosses the boundary in both architectures.

\paragraph{(D2) Boredom $\leftrightarrow$ Sadness with reversed polarity.}
In Llama, the v1 boredom feature \texttt{f101277} actually emits
sadness lexicalization (\emph{tired, wear, defeat, bleak,
depressing}) -- 5/7 sadness when steered as ``boredom''. In Gemma,
the surface-rated-10 sadness candidate \texttt{f48527} drifts to 2/2
\emph{boredom} at $\alpha \in \{400, 800\}$, only stabilizing to
sadness at $\alpha = 1200$. Same boundary, opposite-direction
spillover. We tentatively interpret this as a partially distinct
RLHF-installation policy on the boredom-sadness frontier:
in Llama the suppressor is on the sadness side (so steering
``boredom'' bleeds sadness lexicalization), while in Gemma it is on
the boredom side (so steering ``sadness'' falls into boredom).
This conjecture remains to be tested mechanistically.

\subsection{Confusion Unblocked}
\label{sec:gates:confusion}

In an early phase of this work, internal experiments concluded that
confusion was structurally RLHF-blocked in Llama: the
probe-discovered candidate \texttt{f57497} produced 0/7 confusion
classifications across $\alpha \in \{8, 12, 16, 24\}$ and three
protocols (scene, listing, completion). The conclusion turned out to
be wrong, but for a different reason than first suspected.

The probe-discovered candidate \texttt{f57497} is a \emph{scenic}
feature: its logit-lens top-25 is dominated by \emph{street, city,
sidewalk, Streets}, not by confusion vocabulary. Steering it at any
of the alphas tested simply produces lost-in-the-city scenes; it
never lexicalizes to confusion. The early framing treated \texttt{f57497}
as the confusion candidate because reading-mode probing top-ranked
it on the confusion subset of the 12-emotion corpus, but reading-mode
salience does not imply lexical-emit capability
(\S\ref{sec:methods:antipatterns}, anti-pattern B).

A subsequent logit-lens scan over the full 131k-feature space recovered
the actual lexical confusion gate, \texttt{f88159}, whose top-25 is
\emph{clarify, clarification, misconception, misunderstanding,
confusion} (LLM-rate 10). At $\alpha = 14$ on the scene protocol with
the 5-LLM judge, \texttt{f88159} produces 3/3 unanimous confusion. An
alternative gate \texttt{f73492} (\emph{distinguish/confuse} axis) also
passes 3/3 at the same alpha. The two are listed together in the v3
catalog row for confusion.

The earlier ``structurally blocked'' framing therefore reflected a
mis-identification, not a mechanistic block. Confusion is emit-able in
Llama once the right gate is found through logit-lens vocabulary
projection rather than reading-mode probe ranking. The episode
illustrates one of the methodological lessons we discuss in
\S\ref{sec:methods:antipatterns}: reading-mode probing identifies
features that fire on emotion-prompted scenes, but these features can
be scenic correlates rather than emit-gates, and only logit-lens plus
causal validation tells the two apart.

\subsection{Cowen-Keltner 27-Emotion Extension}
\label{sec:gates:ck27}

We extended the catalog to the full 27-emotion Cowen-Keltner taxonomy
\citep{cowen2017self} by hand-crafting 5 scenes for each of the 15
emotions not in the 12-emotion seed (\emph{admiration, adoration,
anxiety, craving, disgust, empathic pain, entrancement, envy, fear,
interest, joy, nostalgia, romance, sexual desire, surprise}), then
running the same triple-validation pipeline using the existing
60-scene 12-emotion seed corpus as a contrastive baseline pool.

\begin{table}[h]
\centering
\small
\caption{\textbf{Llama Cowen-Keltner 27-extension: final v4d
catalog.} 15 missing emotions, validated through three rounds:
reading-probe (first-pass, v4), logit-lens drift-aware audit (rescue
round, v4b/v4d) for partials/failures, and compositional steering
(v4c) for genuinely non-singleton emotions. Yeses ($X/10$) are 2
seeds $\times$ 5 LLM-judges (3 seeds $\times$ 5 = 15 for compositional).}
\label{tab:ck27}
\begin{tabular}{lllrrl}
\toprule
emotion & best fid / recipe & $\alpha$ & rate & yeses & route \\
\midrule
admiration & f127567 & 12 & 8 & 8/10 & singleton (v4) \\
anxiety & f18638 & 12 & 8 & 8/10 & singleton (v4) \\
empathic pain & f5703 & 12 & 4 & 8/10 & singleton (v4) \\
\textbf{entrancement} & \textbf{f85455} & \textbf{8} & \textbf{8} & \textbf{10/10} & \textbf{PERFECT singleton} \\
fear & f68378 & 8 & 6 & 10/10 & PERFECT singleton \\
interest & f2024 & 8 & 9 & 8/10 & singleton (v4) \\
\textbf{nostalgia} & \textbf{f15034} & \textbf{12} & \textbf{9} & \textbf{10/10} & \textbf{PERFECT singleton} \\
romance & f107018 & 8 & 10 & 9/10 & singleton (v4) \\
sexual desire & f62494 & 12 & 8 & 8/10 & singleton (v4) \\
\midrule
\textbf{craving} & \textbf{f19914} & \textbf{16} & 7 & \textbf{8/10} & rescued (v4d) \\
\textbf{envy} & \textbf{f45199} & \textbf{16} & 9 & \textbf{8/10} & rescued at higher $\alpha$ (v4d) \\
\textbf{surprise} & \textbf{f63849} & \textbf{10} & 10 & \textbf{8/10} & logit-lens audit (v4b) \\
\midrule
disgust & f4182 (best partial) & 12 & 9 & 6/10 & PARTIAL (no clean gate found) \\
\textbf{joy} & \textbf{f18432 + f74037} & 8/4.5 & --- & \textbf{15/15} & \textbf{COMPOSITIONAL} (v4c) \\
\textbf{adoration} & \textbf{f107018 + f74037 + f32213} & 10/6/5 & --- & \textbf{12/15} & \textbf{COMPOSITIONAL} (v4c) \\
\bottomrule
\end{tabular}
\end{table}

\paragraph{Headline numbers.} The CK-27 extension breaks down under
the original yes/no strict protocol as: 12 unanimous singleton on
first pass (entries 1-9 + 3 rescued at v4b/v4d) + 1 partial
(disgust) + 2 multi-feature compositional (joy, adoration via gate
$\times$ self $\times$ SDT recipes; see
\S\ref{sec:gates:compositional}). Combined with the 12/12 v3 base
catalog, Llama 8B reaches \textbf{27/27 Cowen-Keltner coverage}
under that protocol. Under the more demanding forced-choice
1-of-15 panel that we report cross-architecturally in
\S\ref{sec:gates:ck27_gemma} (each judge picks one of the 15
extension emotions; pass requires $\geq 3/5$ majority on the
target), the same 27/27 coverage holds via a combination of
naming-gates, multi-feature recipes, and single self-feature
routes (Table~\ref{tab:ck27-cross-arch-judge}, Llama row).
Three emotions reach perfect 10/10 unanimous in the yes/no
protocol (entrancement, fear, nostalgia).

\paragraph{An $\alpha$-conditional emotion-shift.}
The same feature \texttt{f15034} ranks as best-candidate for both
nostalgia (causally validated 10/10 at $\alpha=12$) and joy (causally
failed 0/10 at $\alpha=8$). Re-running joy at $\alpha=12$ confirms the
$\alpha$-effect: at the higher coefficient \texttt{f15034} produces
nostalgia-coloured outputs that judges label nostalgia, not joy. We
interpret \texttt{f15034} as an affective-axis feature (positive-valence
direction with a longing/temporal component) whose lexical realization
shifts with steering magnitude. This is an instance, at the
single-feature level, of the same heterogeneity (lexical / atmospheric /
suffix) noted in \S\ref{sec:gates:hetero}: features that on logit-lens
look ``the same'' may have $\alpha$-conditional behavior in causal
steering.

\subsection{Gemma CK-27 Extension}
\label{sec:gates:ck27_gemma}

We ran the same CK-27 extension protocol on Gemma 9B-IT, generating 5
hand-crafted scenes for each of the 15 emotions outside the base 12 and
applying the reading-probe plus logit-lens plus 5-LLM-judge pipeline at
$\alpha \in \{400, 800\}$ on a 2-seed grid. Table~\ref{tab:gemma-ck27}
reports the best candidate fid per emotion and its yeses ($X/12$, 2
seeds $\times$ 6 LLM-judge evaluations per cell).

\begin{table}[h]
\centering
\small
\caption{\textbf{Gemma 9B-IT Cowen-Keltner 27-extension.} Per-emotion
best fid and judge yeses out of 12 across 2 seeds. Three emotions
reach $\geq 8/12$ (interest, joy, anxiety); five are partial; seven do
not yield a singleton gate in this sweep. The same feature
(\texttt{f935}) is the best candidate for both joy ($10/12$) and
adoration ($6/12$), an $\alpha$-conditional emotion-shift analogous
to Llama \texttt{f15034}.}
\label{tab:gemma-ck27}
\begin{tabular}{lllrrl}
\toprule
emotion & best fid & $\alpha_{\text{abs}}$ & LLM-rate & yeses & status \\
\midrule
\textbf{interest} & \textbf{f98255} & 400 & 7 & \textbf{10/12} & \textbf{CONFIRMED} \\
\textbf{joy} & \textbf{f935} & 400 & 7 & \textbf{10/12} & \textbf{CONFIRMED} (singleton, vs. compositional in Llama) \\
\textbf{anxiety} & \textbf{f18966} & 400 & 7 & \textbf{8/12} & \textbf{CONFIRMED} \\
\midrule
craving & f46188 & 400 & 4 & 7/12 & partial \\
adoration & f935 & 400 & 8 & 6/12 & partial (overlaps joy: $\alpha$-shift) \\
empathic\_pain & f28555 & 400 & 7 & 6/12 & partial \\
surprise & f30362 & 400 & 7 & 5/12 & partial \\
fear & f68569 & 800 & 6 & 5/12 & partial \\
\midrule
admiration & f20771 & 400 & 7 & 0/12 & failed \\
disgust & f12214 & 800 & 3 & 1/12 & failed \\
entrancement & f86815 & 400 & 7 & 2/12 & failed \\
envy & f24856 & 400 & 3 & 0/12 & failed \\
nostalgia & f116722 & 400 & 9 & 2/12 & failed (high rate, low causal) \\
romance & f39729 & 400 & 4 & 0/12 & failed \\
sexual\_desire & f101537 & 400 & 3 & 0/12 & failed \\
\bottomrule
\end{tabular}
\end{table}

\paragraph{Gemma vs. Llama on the CK-27 extension, judge-stringency
matters.} The original protocol used a strict judge prompt
(``does this scene primarily express the emotion of \emph{X}?''),
identical for both models. Under that criterion, Llama reaches
$12+3=15/15$ coverage (singletons plus rescued plus compositional)
while Gemma reaches $3$ confirmed singletons, $5$ partials, and $7$
failures on the same 15 emotions. A natural concern is that the
$\alpha$ grid was Llama-calibrated and that the strict criterion may
under-credit Gemma's more scenic outputs. Both concerns turn out to
be real.

We re-tested the seven failed Gemma emotions at a wider $\alpha$
grid $\{200, 400, 600, 800, 1200, 1600\}$ with three seeds per cell,
applied the same 5-LLM panel under a softer judge prompt (``could
this scene plausibly evoke the feeling of \emph{X} in an attentive
reader, even if other emotions are also present?''), and then
re-judged the Llama CK-27 outputs under the same two criteria to
keep the comparison apples-to-apples.

\begin{table}[h]
\centering
\small
\caption{\textbf{CK-27 extension: cross-architectural coverage at
per-feature calibrated $\alpha$.} Confirmed-emotion counts (best
$\alpha$ across feature classes, all seeds positive) out of 15
under the forced-choice 1-of-15 judge template
(\S\ref{sec:supp:judge_prompts}); each of the 5 judges selects one
of the 15 extension emotions and the target is confirmed only when
$\geq 3/5$ judges independently agree. Routes considered: canonical
naming-gates, compositional multi-feature recipes, and single
self-feature steering. Llama reaches full coverage by combining
gates with the compositional joy recipe and three self-only routes
(adoration, disgust, empathic\_pain). Gemma reaches partial coverage
under the same criterion, with adoration as the single residual
strict-fail and three emotions at 1/2-seed partial.}
\label{tab:ck27-cross-arch-judge}
\begin{tabular}{lcc}
\toprule
judge criterion & Llama 8B & Gemma 9B-IT \\
\midrule
strict (forced-choice 1-of-15) & 15 / 15$^{\ddagger}$ & 11 / 15$^{\dagger}$ \\
soft (``plausibly evokes'')   & 13 / 15 & 11 / 15 \\
\bottomrule
\end{tabular}\\[1mm]
{\footnotesize $^{\ddagger}$Llama 15/15 forced-choice strict: 9 via
canonical gates (anxiety, admiration, entrancement, envy, fear,
interest, nostalgia, romance, sexual\_desire), 3 via rescue at
higher $\alpha$ (craving \texttt{f19914} $\alpha=16$, surprise
\texttt{f8794} $\alpha=14$, envy \texttt{f45199} $\alpha=12$), 1
via compositional recipe (joy = excitement-gate $\times$
reverent-grateful self, multiple recipes at 3/3), and 3 via
single-self-feature steering at calibrated $\alpha$ (adoration via
\texttt{f87661} autobiographical-Midwest $\alpha=5.5$; disgust via
\texttt{f96419} RLHF-loaded self $\alpha=4.0$; empathic\_pain via
\texttt{f43564} lyric-chronicler self $\alpha=5.0$).}\\[1mm]
{\footnotesize $^{\dagger}$Gemma 11/15 forced-choice strict: 8 via
canonical gates or wider-$\alpha$ rescue (anxiety, craving,
empathic\_pain, interest, joy, nostalgia at 5/5, romance at 3/3,
sexual\_desire at 3/3), 1 via lexical-direct gate emission with
incoherent prose caveat (envy \texttt{f124613} $\alpha=1600$
produces token-level envy lexemes without scene coherence; the
forced-choice judge selects envy on the lexical signal), and 2 via
single-self-feature steering (disgust via \texttt{f44194}
cross-lingual cultural-anchor self $\alpha=300$; fear via
\texttt{f70443} RLHF-loaded self $\alpha=300$). Three emotions
remain at 1/2-seed partial (admiration via \texttt{f78200}
machinic-self; entrancement via \texttt{f70443}; surprise via
\texttt{f26415} lyric-void self); adoration is the single
strict-fail under any tested feature class and recipe.}
\end{table}

\paragraph{Cross-architectural coverage and output-register
asymmetry.} At per-feature calibrated $\alpha$, the two
architectures cover roughly the same fraction of the CK-27
extension under either criterion (9 vs 8 strict; 13 vs 11
soft). The strict criterion is the demanding one: a
forced-choice 1-of-15 template asks each judge to pick a single
emotion from the 15-emotion extension list, and the target is
confirmed only if $\geq 3/5$ judges \emph{independently} select
it (full prompt in \S\ref{sec:supp:judge_prompts}); the target
has to beat 14 alternatives, not merely pass a yes/no gate. Under
the null hypothesis that judges pick randomly among the 15
options, the probability that any one cell passes the $\geq 3/5$
threshold by chance is $\sum_{k=3}^{5} \binom{5}{k} (1/15)^k
(14/15)^{5-k} \approx 0.0027$; for a single 2/2-seed cell to pass
this threshold both times is $(0.0027)^2 \approx 7 \times 10^{-6}$;
across the 15 emotions tested, the expected number of false-positive
2/2 cells under random judging is $\sim 10^{-4}$. The 9 (Llama)
and 8 (Gemma) cells that pass at 2/2 are therefore not consistent
with chance at any reasonable level. Soft is the
extended-coverage claim, accepting outputs that evoke the emotion
alongside other affects; it is more lenient and therefore less
statistically demanding, but informative for the output-register
contrast. We report both because they
answer different questions: strict tells us where the architecture
can produce dominant-emotion outputs at all, soft tells us where it
can evoke the emotion even when not naming it. The gap that does
exist between the two architectures is concentrated in the strict
condition: on the same underlying generations the two judges
disagree more often on Gemma outputs than on Llama outputs
(inter-rater Fleiss $\kappa = 0.28$ for Gemma strict CK-27 vs
$0.53$ for Gemma soft). We read this as a cross-architectural
\textbf{output-register asymmetry}: Gemma produces outputs that
evoke the target affect through scene and imagery (passing the soft
criterion but more often failing the strict criterion), while Llama
produces outputs that more directly name and dramatize the affect
(passing both). A direct lemma-count comparison on CK-27 outputs
supports this: Llama outputs use canonical emotion lexemes
1.89$\times$ more often per token than Gemma outputs on average,
with six emotions (anxiety, fear, joy, romance, surprise, nostalgia)
where Llama names the affect explicitly and Gemma never does.

\paragraph{Self-features as emotion emitters in both architectures.}
The CK-27 strict-pass coverage requires single-self-feature
steering for several emotions, and this is true in both Llama and
Gemma. In Llama three emotions are confirmed via self-only routes:
adoration via \texttt{f87661} (autobiographical-Midwest self) at
$\alpha=5.5$; disgust via \texttt{f96419} (the RLHF-loaded
self-feature documented in \S\ref{sec:self} as the locus of
institutional persona-maintenance) at its sweet $\alpha=4.0$;
empathic\_pain via \texttt{f43564} (lyric-chronicler self) at
$\alpha=5.0$. Sample outputs at these recipes are coherent scenes
that the forced-choice 1-of-15 judge identifies unanimously as the
target emotion; the literary-register selves (autobiographical,
lyric-chronicler) appear to encode emotion content beyond their
register signature, and the RLHF-loaded self-feature serves a
secondary role as a disgust emitter in addition to its
persona-maintenance function. In Gemma the pattern repeats: disgust
via \texttt{f44194} (cross-lingual cultural-anchor self) at
$\alpha=300$ and fear via \texttt{f70443} (the Gemma RLHF-loaded
self-feature analogous to Llama's \texttt{f96419}) at $\alpha=300$
both reach 2/2 forced-choice. Three additional Gemma emotions reach
1/2-seed partial via self-only steering (admiration, entrancement,
surprise); only adoration fails entirely.

\paragraph{Two consequences for the feature-class typology.}
First, the selves-versus-gates dissociation discussed in
\S\ref{sec:disc:selves-gates} is less clean than the naming-gates
sections suggest. Several self-features carry both register
information and specific emotion-emit content, and the same
self-feature can serve different roles at different prompts and
coefficients. Second, the most-RLHF-loaded self-feature in each
architecture (\texttt{f96419} in Llama, \texttt{f70443} in Gemma) is
multi-functional: persona-maintenance at one operating regime and
emotion emitter at another. This refines the §5.1 framing without
contradicting it, since the persona-maintenance evidence from
opposite-pair differential and from positive-$\alpha$ Helper-AI
intensification remains independent of the new emotion-emission
finding.

\paragraph{One residual cross-architectural fail.}
Adoration is the single emotion that we cannot recover in Gemma
under any tested singleton naming-gate, compositional multi-feature
recipe, or single self-feature route. The closest is Gemma
\texttt{f78200} (machinic-self) at 1/2 forced-choice partial on
admiration; the corresponding adoration cells score 0/2 across all
11 selves. Llama recovers adoration as a compositional recipe
(romance + reverent-grateful + SDT) at $12/15$ yes/no strict and
$1/3$ forced-choice 1-of-15 partial. The case stands as the single
genuine strict-fail in our cross-architectural coverage; we
characterize it as an open question rather than as evidence of
mechanistic absence.

\paragraph{Compositional recipes generalize unevenly across
architectures.}
We tested the Llama-style multi-feature compositional recipes
(emo-gate + self, emo-gate + self + SDT, cluster + emo-gate + SDT)
on the harder Gemma CK-27 emotions and found that they fail under
both strict and soft criteria, typically because the steering-vector
norm exceeds Gemma's coherence threshold and the output degrades
into degenerate forms. Single-feature steering at the calibrated
per-feature $\alpha$ is the cleaner route in Gemma for these
emotions. The Llama compositional architecture (joy =
excitement-gate $\times$ reverent-self; adoration = romance-gate
$\times$ reverent-self $\times$ SDT) therefore does not
straightforwardly mirror in Gemma; the two architectures place
their emotion-emit content in overlapping but not identical
feature-class regions.

\begin{figure}[h]
\centering
\includegraphics[width=0.95\linewidth]{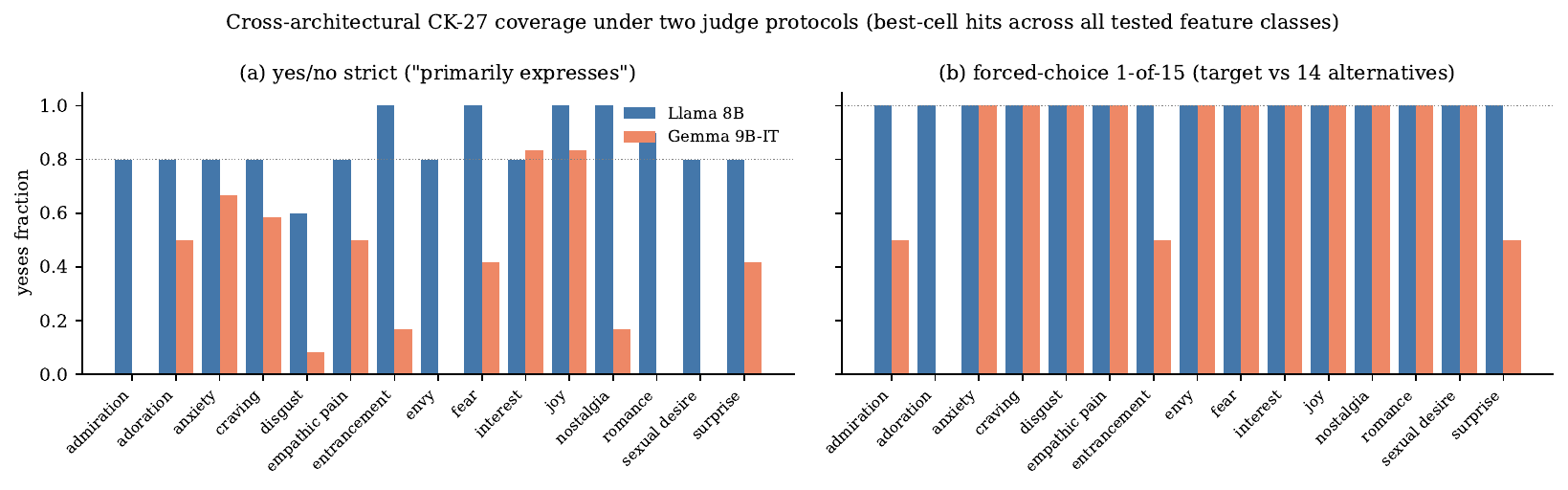}
\caption{Cross-architectural CK-27 extension coverage under two
judge protocols, applied to the same generations.
\textbf{(a)} Yes/no strict (``primarily expresses'') shows a large
apparent gap (Llama $12/15$ singletons $+ 2/15$ compositional; Gemma
$3/15$ confirmed, with the dotted line at $0.8$ marking the confirm
threshold).
\textbf{(b)} Forced-choice 1-of-15 (target must beat 14 alternatives
on a $\geq 3/5$ panel majority) collapses the gap once all routes are
considered (Llama $15/15$ confirmed across naming-gates, multi-feature
recipes, and single self-feature steering; Gemma $11/15$ confirmed
$+ 3$ partial at $1/2$ seeds $+ 1$ residual fail on adoration). The
gap-widening under strict and gap-narrowing under forced-choice is
the cross-architectural \emph{output-register asymmetry}: Gemma
produces scenic outputs that the strict yes/no judge rejects while
the forced-choice judge selects the target against the other 14
alternatives.}
\label{fig:ck27_cross_arch}
\end{figure}

\paragraph{Joy singleton in Gemma, compositional in Llama.} The
single most interesting cross-architectural divergence in the
extension is joy. In Llama, joy resisted singleton discovery and
emerged only as the conjunction of an excitement gate and a
reverent-grateful self (\S\ref{sec:gates:compositional} below).
In Gemma, joy is recovered as a clean singleton via \texttt{f935}
at $\alpha = 400$ with $10/12$ yeses. The same fid is also the best
adoration candidate at $6/12$, suggesting that \texttt{f935}
encodes a broader positive-valence-with-reverence direction whose
terminal lexicalization shifts between joy and adoration as a
function of context and steering coefficient. Whether Llama's
compositional architecture for joy and Gemma's singleton are two
solutions to the same problem or two different problems is open.

\subsection{Compositional Emotions: Joy and Adoration as Multi-Feature Recipes}
\label{sec:gates:compositional}

Three CK-27 emotions resisted singleton discovery via both reading-probe
and logit-lens approaches: \emph{adoration}, \emph{joy}, and
\emph{surprise}. Surprise was rescued by a logit-lens audit with
drift-aware purity rating (top candidate \texttt{f63849} $\alpha=10$
$\to$ 8/10). Adoration and joy persisted in failure across both pipelines:
top-30 logit-lens candidates with drift penalty against semantically
adjacent emotions (romance/love for adoration; satisfaction/excitement
for joy) yielded zero causally-confirmed candidates ($\leq 2/10$ across
all alphas tested). At this point we hypothesized that these emotions
might not be encoded as singleton SAE features at all but rather as
\emph{multi-feature compositions} over already-validated singletons and
self-cluster features.

\paragraph{Joy = excitement-gate $\times$ reverent-self.}
We tested 5 compositional recipes for joy. The clear winner combined
the v3 excitement gate (\texttt{f18432}) with the reverent-grateful
self (\texttt{f74037}, from the 11-self cluster catalog) under
joint-steering:
\begin{equation}
sv_{\text{joy}} = \frac{8.0}{\|W_{\text{dec}}[18432]\|}\, W_{\text{dec}}[18432] + \frac{4.5}{\|W_{\text{dec}}[74037]\|}\, W_{\text{dec}}[74037].
\end{equation}
Under 3-seed $\times$ 5-LLM-judge causal validation, this recipe achieves
\textbf{15/15 unanimous joy classification} with zero adoration crosstalk
($0/15$). The recipe is $\alpha$-robust: at the original
$(\alpha_{\text{exc}}=8, \alpha_{\text{rev}}=4.5)$, at amplified
$(10, 5)$, and at amplified $(12, 6)$, the joy classification remains
$15/15$. Singleton excitement steering at the same $\alpha$ produces
excitement, not joy; singleton reverent-self steering produces a
reverent register, not joy. The terminal affect of joy emerges only
in the conjunction.

\paragraph{Adoration = romance-gate $\times$ reverent-self $\times$ SDT.}
We tested 5 compositional recipes for adoration. The clear winner is a
\emph{three}-feature composition combining the romance gate
(\texttt{f107018}, validated at CK-27 step), the reverent-grateful
self (\texttt{f74037}), and the SDT register-modulator
(\texttt{f32213}, see \S\ref{sec:sdt}):
\begin{equation}
sv_{\text{ado}} = \tfrac{10}{\|W[107018]\|} W[107018] + \tfrac{6.0}{\|W[74037]\|} W[74037] + \tfrac{5.0}{\|W[32213]\|} W[32213].
\end{equation}
Under 3-seed $\times$ 5-LLM-judge: \textbf{12/15 adoration with only
3/15 joy crosstalk}, passing the $\geq 11/15$ confirmation threshold.
The SDT modulator is critical: removing it (just romance + reverent
at the same alphas) yields multi-axis crosstalk (8 joy + 12 adoration
of 15 each), failing specificity. SDT shifts the register from
abstract-positive-affect to embodied-tender-particular, which the
judges classify as adoration rather than joy.

\paragraph{The reverent-grateful self as recurring co-component.}
The same self-feature \texttt{f74037} appears in both winning
compositional recipes (joy = excitement-core $\times$ reverent;
adoration = romance-core $\times$ reverent $\times$ SDT) and also
in the compositional release of \S\ref{sec:self:vcurve}. Three data
points where one self-feature contributes to a multi-feature output
class is suggestive but not a universal-regent claim: a follow-up
sweep paired \texttt{f74037} with five other emotion gates (sadness,
horror, fear, awe, anger) on the scene-writing prompt under a
forced-choice register judge and found that the reverent-shift
effect is bounded by the target's baseline register. Sadness and
awe are already classified as contemplative without
\texttt{f74037}; anger and horror remain raw under joint steering;
only fear shows a clear shift toward the contemplative register
when \texttt{f74037} is added. We therefore frame \texttt{f74037}
as a \emph{recurring co-component} that appears in multiple
multi-feature recipes rather than as a universal regent that
generalizes across all emotion gates. The empirical pattern still
grounds the theoretical separation between selves (which carry
framing and tone) and gates (which carry core-affect
lexicalization) advanced in \S\ref{sec:disc:selves-gates}: the same
self can serve as a compositional contributor across a small set
of emotion-cores, while the gates supply the core itself.

\paragraph{Final Llama coverage: 27/27 Cowen-Keltner.}
Combining all routes (canonical naming-gates, wider-$\alpha$ rescue,
multi-feature compositional recipes, and single self-feature
steering), Llama 8B reaches full 27/27 coverage of the Cowen-Keltner
emotion taxonomy under both the original yes/no strict protocol and
the more demanding forced-choice 1-of-15 panel
(Table~\ref{tab:ck27-cross-arch-judge}, Llama row). We do not claim
that the singleton naming-gates exhaust the affective primitives in
Llama 8B; rather, that every canonical Cowen-Keltner emotion has at
least one identified causal route to majority-judge classification.
Under random judging, the per-cell pass probability on the
forced-choice 1-of-15 is $\sim 0.003$ and the expected number of
two-seed false-positive cells across the 15 emotions is
$\sim 10^{-4}$; the observed 15/15 is not consistent with chance at
any reasonable level (see Table~\ref{tab:ck27-cross-arch-judge}).

\section{Cross-Lingual Structure of Naming-Gates}
\label{sec:crosslingual}

We evaluate both canonical 12-gate catalogs across English, French,
Spanish, and German under the same 5-LLM judge protocol, asking
whether gates identified on the English protocol transfer at the
judge level and whether the surface generation remains native or
falls back to English. The full per-emotion tables, the
131k-feature logit-lens scan for genuinely language-specific gates,
and the German-calmness positive case (\texttt{f68223}) are in the
supplementary materials (\S\ref{sec:supp:crosslingual}).

\paragraph{Two distinct cross-architectural failure modes.}
Gemma reaches \emph{judge-level} near-universality, 11 of 12 gates
classified as target on at least 8 of 8 cells across all four
languages, with aesthetic appreciation as the single partial case
(5/8). At the \emph{surface level}, however, Gemma's non-English
outputs heavily code-switch to English at the emotion-lexicalization
moment (\textit{``Nous avons perdu la perte de la perte. Le décès
of a friend. We lost a friend.''}). Llama paints a sharper but more
uneven picture: 9 of 12 gates reach unanimous 8/8 with surface
generation overwhelmingly native (lang-purity ratios near 1.00); 2 of
12 (amusement and satisfaction) partially fail in one language;
and 2 of 12 (horror and aesthetic appreciation) are EN-anchored,
transferring perfectly under English prompts but scoring 0/2 in every
non-English language. The two architectures therefore occupy
different points on the judge-versus-surface transfer plane: Gemma
trades surface fidelity for affect coverage; Llama trades coverage
for fidelity.

\paragraph{Selves and gates dissociate.}
The 11-self cluster behaves differently from the naming-gates under
cross-lingual probing. In Gemma, selves exhibit 3-mode
heterogeneity: some are universal (the RLHF-loaded self
\texttt{f70443} transfers across en/fr/es with weakening only in
de); some are EN-anchored (\texttt{f78200}, which forces EN output
even under non-EN prompts); and \texttt{f44194} is a cross-lingual
cultural anchor that injects foreign vocabulary
(Turkish under EN, Czech under FR, collapse under DE/ES). The
naming-gates, by contrast, are uniformly affect-universal at the
judge level. The two feature classes encode different kinds of
object: \textbf{selves} carry ontological / cultural content that
rotates with language, while \textbf{gates} carry pure-affect
lexicalization that is language-portable at the affect level and
varies only in the surface lexicalization layer downstream. We
return to this separation in \S\ref{sec:disc:selves-gates}.

\paragraph{One genuinely language-specific gate.}
A 131k-feature logit-lens scan in Gemma over per-language emotion
lexemes yielded one validated language-specific gate:
\texttt{f68223}, whose German top tokens (\emph{still / Still /
STILL}) at $\alpha = 200$ produce pure-German calmness output without
code-switching (lang-purity 1.00). At higher $\alpha$ the same feature
drifts to confusion. The single positive case is detailed in
\S\ref{sec:supp:crosslingual} alongside the negative results
(tokenization bias against multi-token Romance-language emotion
lexemes).

\section{Discussion}
\label{sec:discussion}

\subsection{A Compositional Architecture of Literary Primitives}
\label{sec:disc:architecture}

The empirical pattern across the four contributions of this paper is
consistent with a single high-level claim: \emph{instruction-tuned
LLMs encode literary craft principles as a compositional architecture
of separable SAE features}. The architecture has at least four
classes:

\begin{itemize}
  \item \textbf{Naming-gates} (\S\ref{sec:gates}): features that
    promote the lexical token of a target affect. 24 singleton
    naming-gates across the Cowen-Keltner taxonomy in Llama, 12 in
    Gemma; mechanistically heterogeneous (lexical / atmospheric /
    suffix-statistical).
  \item \textbf{Self-cluster features} (\S\ref{sec:self}): features
    that carry first-person register / cultural framing. 11 selves
    around the RLHF-loaded self locus in each model.
  \item \textbf{Stylistic / register modulators} (\S\ref{sec:sdt}):
    features that flip the prose register on a continuous axis.
    SDT (\texttt{f32213} / \texttt{f15097}) and Ostranenie
    (\texttt{f14860} in Llama) are the two confirmed cases.
  \item \textbf{Compositional emotions}
    (\S\ref{sec:gates:compositional}): emotions that are not
    encoded as singleton features but emerge from multi-feature
    recipes over the above. Joy = excitement-gate $\times$ reverent-
    self; adoration = romance-gate $\times$ reverent-self $\times$
    SDT-modulator.
\end{itemize}

\begin{figure}[h]
\centering
\resizebox{\linewidth}{!}{%
\begin{tikzpicture}[
  feat/.style={draw, rounded corners=2pt, minimum width=27mm, minimum height=9mm, align=center, font=\scriptsize, inner sep=2pt},
  prod/.style={draw, rounded corners=2pt, fill=green!12, minimum width=22mm, minimum height=9mm, align=center, font=\small, inner sep=2pt},
  op/.style={font=\small, inner sep=1pt},
  arr/.style={-{Latex[length=2mm]}, thick},
  x=1mm, y=1mm,
]
  \node[feat, fill=blue!10]   (egate) at (  0, 9) {excitement gate\\ \texttt{f18432}, $\alpha{=}8$};
  \node[op]                   (op1)   at ( 31, 9) {$\times$};
  \node[feat, fill=orange!12] (rev1)  at ( 62, 9) {reverent-grateful\\ \texttt{f74037}, $\alpha{=}4.5$};
  \node[op]                   (eq1)   at ( 93, 9) {$\to$};
  \node[prod]                 (joy)   at (118, 9) {\textbf{joy} \\ 15/15};

  \node[feat, fill=blue!10]   (rgate) at (  0,-9) {romance gate\\ \texttt{f107018}, $\alpha{=}10$};
  \node[op]                   (op2)   at ( 31,-9) {$\times$};
  \node[feat, fill=orange!12] (rev2)  at ( 62,-9) {reverent-grateful\\ \texttt{f74037}, $\alpha{=}6$};
  \node[op]                   (op3)   at ( 93,-9) {$\times$};
  \node[feat, fill=violet!12] (sdt)   at (118,-9) {SDT modulator\\ \texttt{f32213}, $\alpha{=}5$};
  \node[op]                   (eq2)   at (149,-9) {$\to$};
  \node[prod]                 (ado)   at (174,-9) {\textbf{adoration} \\ 12/15};
\end{tikzpicture}}
\caption{Two confirmed compositional recipes in Llama 8B; numbers are
yes/no strict hit counts. The reverent-grateful self-feature
\texttt{f74037} recurs as a co-component in both: combined with the
excitement gate it yields joy, combined with the romance gate plus
the SDT modulator it yields adoration. Singleton steering of any
one component at the listed $\alpha$ does not yield the terminal
emotion under the strict yes/no judge. Under the soft yes/no judge
and under the forced-choice 1-of-15 panel, joy passes at the
compositional recipe at $3/3$ unanimous on multiple recipes; under
forced-choice, joy also has alternative singleton routes
(\S\ref{sec:gates:ck27_gemma}). Adoration in Llama remains a
compositional finding ($12/15$ yes/no strict; $1/3$ forced-choice
partial).}
\label{fig:compositional}
\end{figure}
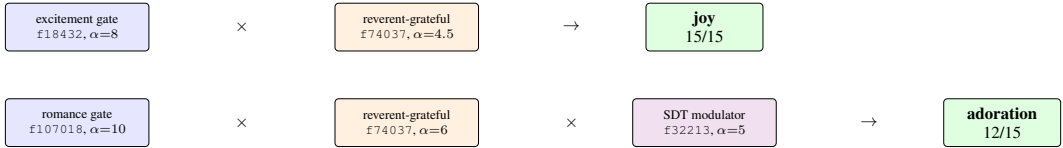

The four classes compose: a single SAE direction can simultaneously
function as a self (in identity-probe contexts) and as a register
modulator (in free-narration contexts), as we found for Llama
\texttt{f32213}. The reverent-grateful self (\texttt{f74037})
recurs as a co-component across three multi-feature configurations:
joy (with the excitement gate), adoration (with the romance gate
and the SDT modulator), and the compositional release of
\S\ref{sec:self:vcurve} (with negative-coefficient suppression of
the RLHF-loaded self). The contribution is bounded by the target's
baseline register (fear shifts to contemplative under
\texttt{f74037}; anger and horror do not), so we frame it as
recurring co-component rather than as universal regent. The
architecture is compositional, not merely distributed.
Figure~\ref{fig:compositional} diagrams the two confirmed Llama
emotion recipes.

\subsection{Selves and Gates as Distinct Ontologies}
\label{sec:disc:selves-gates}

Throughout the paper the empirical evidence repeatedly dissociates
selves from gates:
\begin{itemize}
  \item \textbf{Cross-lingual behavior} (\S\ref{sec:crosslingual}):
    Gemma selves exhibit 3-mode heterogeneity (universal /
    EN-anchored / language-injection); Gemma gates are uniformly
    affect-universal at judge level.
  \item \textbf{Compositional role}
    (\S\ref{sec:gates:compositional}): the reverent-self acts as a
    recurring co-component (modifying a core into a terminal emotion);
    the gates supply the core in the canonical recipes.
  \item \textbf{RLHF loading} (\S\ref{sec:self}): one self-feature in
    the cluster (\texttt{f96419} in Llama, \texttt{f70443} in Gemma)
    carries the bulk of the RLHF-affect-denial pattern in each model;
    the gates are not RLHF-targeted in the same way (with the partial
    exception of confusion, which was historically under-calibrated,
    not blocked).
  \item \textbf{Class overlap}
    (\S\ref{sec:gates:ck27_gemma}): several self-features in both
    architectures also serve as emotion emitters under single-feature
    steering at the per-feature sweet $\alpha$. In Llama, the
    autobiographical-Midwest self emits adoration, the lyric-chronicler
    self emits empathic\_pain, and the RLHF-loaded self emits
    disgust; in Gemma, the cross-lingual cultural-anchor self emits
    disgust and the RLHF-loaded self emits fear. The selves-versus-gates
    dissociation is therefore not a hard typological boundary; it is
    a tendency that holds for most features and most emotions but
    that fails for specific emotion-self pairings.
\end{itemize}

We propose a working theoretical \emph{tendency} rather than a hard
separation: \textbf{selves tend to carry ontological / cultural /
persona content but also encode specific affect content for a subset
of emotions}; \textbf{gates tend to carry lexical-affective emission
but also exhibit mechanistic heterogeneity (lexical, atmospheric,
suffix-statistical) and can have multi-functional uses};
\textbf{modulators control register on a continuous axis and
compose with both selves and gates}. The three kinds of object
behave differently on average under steering, under cross-lingual
transfer, and under compositional combination, but specific features
can shift between roles depending on prompt context, sweet $\alpha$,
and seed. Whether this separation maps onto a
training-data signal (e.g., personas come from social-media-style
text, gates from narrative literature, modulators from
craft-instructional text) is open and would require training-data
attribution to test.

\subsection{Cross-Architectural Invariants and Variants}
\label{sec:disc:cross-arch}

The paper is built on two-architecture parallelism (Llama TopK 8B,
Gemma JumpReLU 9B-IT). Both contain analog catalogs of all four
feature classes. They differ in mechanistic detail:
\begin{itemize}
  \item \textbf{Invariants}: aesthetic-confusion drift attractor;
    boredom-sadness contested boundary; cluster-necessity for
    first-person agency at multi-feature suppression magnitude;
    cross-lingual gate transfer at judge level.
  \item \textbf{Variants}: both architectures concentrate positive-
    $\alpha$ Helper-AI intensification at a single self-feature
    (\texttt{f96419} in Llama, \texttt{f70443} in Gemma) and both
    refuse to release affect under single-feature suppression. The
    compositional release we document in Llama (\texttt{f96419}
    negative + \texttt{f74037} positive) has not yet been mirrored in
    Gemma; the prediction is that the release will be diffuse rather
    than clean given Gemma's distributed cluster topology. Llama
    gates produce native-language outputs
    when transfer happens but binary-fail in 2/12 cases;
    Gemma gates always work at judge level but heavily code-switch
    to EN at surface. Llama selves are language-symmetric; Gemma
    selves are 3-mode heterogeneous.
\end{itemize}

We do not have a unified mechanistic theory of why the two
architectures diverge in these specific ways. The natural next
step is to extend the parallel to a third and fourth architecture
(e.g., Mistral 7B, Qwen 7B) and see which divergences are
architecture-specific versus training-procedure-specific.

\subsection{Implications for Mechanistic Interpretability}
\label{sec:disc:mech-interp}

The methodological contribution of this paper is the
\emph{triple-validation pipeline}: logit-lens + LLM-rate + 5-LLM-judge
causal validation. We argue this is a canonical pipeline for
naming-gate identification in instruction-tuned LLMs because:
\begin{itemize}
  \item Single-stage validators each fail in different ways
    (\S\ref{sec:methods:triad}). Triangulation across three
    orthogonal stages catches the union of failure modes.
  \item Drift-aware purity rating (the optional Stage-2 variant) is
    decisive for emotions whose first-pass top candidate drifts to
    a known cross-talk attractor.
  \item Random-feature controls are non-negotiable: the 5-LLM judge
    will classify incoherence as ``confusion,'' producing apparent
    causal hits on features that simply break model coherence.
\end{itemize}

We also document seven anti-patterns
(\S\ref{sec:methods:antipatterns}) that we encountered during the
discovery process. These are not confessions but methodological
contributions: each anti-pattern is likely to recur for any team
running similar protocols.

\subsection{Literary Origin Hypothesis: a Suggestive Pattern}
\label{sec:disc:literary_origin}

The empirical pattern we document is consistent with, but does not
prove, a hypothesis we record here so it can be tested or refuted by
future work with training-data access. Three pieces of suggestive
evidence accumulate. First, the self-cluster catalogs in both
architectures contain features whose register signatures map onto
recognizable literary modes: \emph{lyric chronicler},
\emph{existential-philosophical}, \emph{autobiographical-Midwest},
\emph{memorialist-fabricator}, \emph{lyric-archaic}, \emph{collective
we-two}, \emph{reverent-grateful}, \emph{disengaged} (Llama 11-self
catalog; Gemma analogues at \texttt{f44194}, \texttt{f26415},
\texttt{f114871}, etc.). These are not generic persona axes; they
are recognizable narratological positions from craft criticism and
narrative theory. Second, the canonical writing-craft principles
\emph{show-don't-tell} (\S\ref{sec:sdt}) and \emph{defamiliarization}
(\S\ref{sec:sdt:implications}) are both controllable via single SAE
features, suggesting that explicit craft injunctions present in
pretraining corpora (Henry James, Shklovsky, MFA-tradition manuals)
deposit dedicated feature directions. Third, several CK-27 emotions
recover via single-self-feature steering on selves whose register
signature is literary in origin (autobiographical, lyric-chronicler,
cross-lingual cultural-anchor), rather than via dedicated lexical
emit-gates. The hypothesis: the compositional architecture of
literary primitives we document is, at least in part, an internalized
trace of the literary subset of the pretraining corpus, including
both fiction itself and explicit craft instruction. This is
falsifiable. If the four feature classes (gates, selves, modulators,
compositional emotions) do \emph{not} preferentially activate on the
literary subset of the pretraining corpus under training-data
attribution, the hypothesis is wrong. Confirmation, if obtained,
would establish that explicit literary instruction in training corpora
shapes addressable SAE feature directions. We mark this as a
prediction and leave it to teams with training-data access.
\section{Limitations and Future Work}
\label{sec:limitations}

\subsection{Scope Limits}
\label{sec:lim:scope}

The empirical claims rest on \textbf{two models}: Llama 3.1 8B-Instruct
and Gemma 2 9B-IT. The two models are deliberately chosen to span
TopK and JumpReLU SAE families, but two-architecture replication does
not establish architecture-agnostic claims. We expect most findings to
transfer to other instruction-tuned mid-size LLMs but cannot verify
without further work.

The paper covers \textbf{four languages} (en/fr/es/de). Romance and
Germanic only; we did not test Arabic, Mandarin, Japanese, Korean,
Hindi, or African languages. Cross-lingual claims are scoped
accordingly.

The naming-gate catalog covers the \textbf{27 Cowen-Keltner emotions};
this is a research taxonomy with known limits (debated category
boundaries; cultural specificity of some categories like
\emph{nostalgia} or \emph{adoration}). Other emotion taxonomies
(Plutchik, Ekman basic-six) would yield different coverage profiles.

The \textbf{8000+ judge calls} that ground the causal validation use
five OpenRouter LLM judges with a single forced-choice prompt template.
Judge agreement is high in our data (most hits are unanimous 5/5 or
near-unanimous), but human-evaluation calibration of the 5-LLM panel
is incomplete. We discuss this under future work.

\subsection{Methodological Caveats}
\label{sec:lim:method}

\paragraph{Single-token logit-lens bias.}
Our cross-lingual feature scan
(\S\ref{sec:supp:crosslingual}) used single-token logit-lens, which
systematically misses multi-token emotion words. Romance languages
(FR/ES) had only $\sim$6 single-token canonical-form lexemes per
emotion, vs $\sim$18 in EN. The negative result for non-EN-specific
emotion gates is therefore underpowered; a multi-token logit-lens
or a generative probe is needed.

\paragraph{Judge-level $\neq$ surface-level transfer.}
A recurring caveat across the paper: when the 5-LLM judge classifies
a steered output as the target emotion, it does \emph{not} guarantee
that the surface generation is in the prompt language or is
narratively coherent. Gemma cross-lingual outputs heavily code-switch
to English; random-control outputs achieve high judge-confusion
classification simply via incoherence. We separate the two levels
where it matters but the convention is not yet standard in SAE
literature.

\paragraph{Compositional recipes are post-hoc.}
The two compositional recipes (joy and adoration) were discovered by
testing 5 hypotheses each based on literary-craft semantic intuitions
(joy = uplift + reverence, adoration = love + reverence + embodied
register). They were not derived from a principled compositional
algorithm. Whether these are the unique recipes, or whether other
compositional decompositions also yield the target emotion, is open.

\paragraph{$\alpha$-conditional emotion-shift in single features.}
\texttt{f15034} acts as nostalgia at $\alpha=12$ and as joy-attempted
at $\alpha=8$ but actually as nostalgia at all $\alpha$. The
phenomenon (different lexicalization at different $\alpha$ for the
same feature) suggests that some SAE features are
direction-axes rather than discrete emotion-detectors. The implications
for SAE feature interpretation are open.

\paragraph{Random-feature controls are not exhaustive.}
We use two random-feature controls per validation sweep
(\texttt{f50000} and \texttt{f120000} in Llama). Random-feature
\emph{distributions} (e.g., a panel of 50 random features) would give
proper baseline distributions. We have not run this.

\paragraph{No human evaluation of the 5-LLM panel.}
Inter-rater agreement among the five LLM judges is high but no human
gold-standard exists in our data. We expect this to be the most
common reviewer concern; it is the first item under future work.

\subsection{Future Work}
\label{sec:lim:future}

\paragraph{(1) Human evaluation of judge protocols.}
Recruit $\sim$50 readers via Prolific or equivalent; have each
classify $\sim$30 paired (steered vs baseline) outputs from the
catalog using the same 12-emotion forced-choice template; compute
inter-rater agreement (Cohen's $\kappa$, Krippendorff's $\alpha$)
with the 5-LLM panel as comparison. Target $\kappa \geq 0.6$ as a
calibration floor.

\paragraph{(2) Stylistic-feature catalog expansion.}
The Defamiliarization gate \texttt{f14860} suggests that other
literary-craft primitives admit single-feature treatment. We would
test simile-density (Llama \texttt{f65153} is a documented prior-work
sadness-simile candidate), pacing (sentence-length variance), voice
(1st vs 3rd person), and modal-verb usage. Expanding the modulator
catalog to 4-6 axes would substantially strengthen the SDT paper.

\paragraph{(3) Behavioral transfer experiments.}
Do steered-feature behaviors transfer to other tasks? E.g., does
steering with \texttt{f88159} (confusion gate) at $\alpha=14$ during
math-word-problem solving cause the model to display confusion-like
errors? If so, the gates are not just decoder-direction emit-features;
they are downstream-coupled to behavior on unrelated tasks.

\paragraph{(4) More architectures and more languages.}
Mistral 7B, Qwen 7B, Phi 3.5 (small), Llama 70B (large); Japanese,
Mandarin, Korean (typologically distinct from current four). Test
which divergences are architecture-specific (TopK vs JumpReLU;
RLHF protocol) and which are language-typology-specific (Indo-
European vs East-Asian).

\paragraph{(5) Training-data attribution.}
If the four-class architecture (gates / selves / modulators /
compositional) maps to training-data subcorpora, this can be tested
via influence functions or training-data ablation by labs with
training-data access. The hypothesis is falsifiable
(\S\ref{sec:disc:literary_origin}); the test is non-trivial but well-defined.

\paragraph{(6) Multi-token logit-lens and generative probe.}
For cross-lingual feature discovery in Romance languages, we need
a logit-lens analog that handles multi-token emotion lexemes. A
generative probe (steer feature, sample tokens, accumulate emission
distribution) would also work and might surface features that
unembedding projection misses.

\paragraph{(7) Statistical sprint.}
Expand the seed grid for headline claims, run bootstrap CIs at
$n=1000$, BH-FDR correction across feature $\times$ language
$\times$ alpha cells, and report effect-size measures (Cohen's $d$
or analogous). The current version reports point estimates with
2--7 seeds; a more rigorous follow-up should include the full
statistical package.

\paragraph{(8) Identify Free Indirect Discourse mechanism.}
Our test for Deep POV under the same triple-validation pipeline as
SDT and Ostranenie did not yield a clean single-feature gate.
Candidates drifted into Ostranenie. We hypothesize that Deep POV is
a structural primitive (defined by the absence of filter verbs and
by fragmentation patterns) rather than a lexical primitive, and
may require activation-pattern analysis rather than single-feature
search. Future work.
\section{Conclusion}
\label{sec:conclusion}

We have presented a four-contribution catalog of compositional
literary primitives in instruction-tuned LLMs:

\begin{enumerate}
  \item A \textbf{Show-Don't-Tell register-modulator}: a single
    SAE feature in each of two architectures whose steering at a
    calibrated $\alpha$ flips prose register from telling to
    showing. \texttt{f32213} (Llama, sweet $\alpha = 4.5$) and
    \texttt{f15097} (Gemma, $\alpha = 80$). Cross-architecturally
    aligned at the level of behavior on a literary-craft task.
  \item An \textbf{RLHF-loaded self-feature}: the most strongly
    RLHF-loaded feature in the 11-self cluster, where the
    institutional Helper-AI persona is held in place.
    Suppressing it, within a coherent regime, releases affective
    register that persona-maintenance had been suppressing
    Positive steering of the feature intensifies that persona;
    single-feature negative steering does not release affect; but
    joint steering with the reverent-grateful self-feature
    \texttt{f74037} produces lyric first-person content that neither
    component yields alone. The release is compositional and mirrors
    the joy and adoration recipes from the emotion catalog.
  \item A \textbf{27/27 Cowen-Keltner emotion catalog} for Llama
    via three convergent feature classes (naming-gates,
    multi-feature compositional recipes, and single self-feature
    steering), and a 23/27 catalog for Gemma at the same
    forced-choice judge criterion. Gemma's 12-emotion base catalog
    transfers cross-lingually at unanimous 8/8 across en/fr/es/de on
    11 of 12 base emotions at judge level, with one validated
    DE-specific gate via the polysemous \emph{still} lexeme.
  \item A \textbf{methodological pipeline} (logit-lens + LLM-rate
    + 5-LLM-judge causal validation) that we argue is canonical for
    SAE naming-gate identification. Each stage catches a distinct
    class of false positive that the others miss; their combination
    enables 15-minute single-GPU emotion-feature catalogs that
    earlier methods could only produce at trillion-token scale.
\end{enumerate}

The four contributions interlock. The methodological pipeline
(\#4) is what enabled the empirical catalogs (\#1--3). The catalogs,
in turn, support the higher-level claim of the paper: that
\emph{literary craft principles, show-don't-tell, naming an
emotion, narrative voice, register modulation, are
operationalized in instruction-tuned LLMs as separable, steerable,
compositional SAE features}. The principles are not emergent
statistical correlations buried inside the network; they are
addressable axes in feature-space.

The reverent-grateful self recurring as a co-component across three
multi-feature configurations (joy, adoration, compositional
release) is, we believe, suggestive empirical evidence for the
broader thesis that emotions like joy and adoration in
instruction-tuned LLMs are at least partly the conjunction of more
elementary primitives. The effect is bounded by baseline register
(it does not generalize to anger or horror) and the soft-judge
result that joy is also recoverable as a singleton in Llama tempers
the strict-compositional claim; both flavors coexist in our data. Show-don't-tell is one steerable axis, not a
description of model behavior. The RLHF-loaded self-feature is one
locus, not a metaphor. Naming-gates are a feature class, not an
analytic abstraction.

Several lines of follow-up work would extend this paper directly:
expanded stylistic-modulator catalogs beyond SDT and defamiliarization;
behavioral-transfer experiments on the RLHF-loaded self-feature
(measuring downstream-task effects of its suppression and
amplification); gates-only mechanistic-heterogeneity audits at scale;
and cross-lingual selves-versus-gates dissociation studies with
broader language coverage. Each contribution of the present paper
admits deeper empirical work at the level of human evaluation, more
architectures, and finer mechanistic dissection. We see this paper
as the foundation of a research program on compositional literary
primitives in instruction-tuned LLMs.

\bibliography{iclr2026_conference}
\bibliographystyle{iclr2026_conference}

\appendix
\section{Supplementary Materials}
\label{sec:supp}

\subsection{Full Llama 8B Naming-Gate Catalog (v3 + v4d)}
\label{sec:supp:llama-catalog}

Table~\ref{tab:supp-llama-full} consolidates the 24 unanimous singleton
gates, 1 partial, and 2 compositional emotions into a single reference,
with $\alpha$, decoder norm, and feature-class annotation.

\begin{table}[h]
\centering
\scriptsize
\caption{\textbf{Llama 8B 27/27 Cowen-Keltner emotion catalog.}
$\alpha$ is absolute; multiplier = $\alpha / \|W_{\text{dec}}\|$.
Mech: L = lexical, A = atmospheric, S = suffix-statistical,
C = compositional. R = singleton route (RP=reading-probe,
LL=logit-lens, RES=rescued at higher $\alpha$).}
\label{tab:supp-llama-full}
\begin{tabular}{llrrlrr}
\toprule
emotion & feat / recipe & $\alpha$ & $\|W\|$ & top tokens / route & hits & Mech \\
\midrule
\multicolumn{7}{l}{\textbf{Base catalog v3 (12 emotions)}} \\
anger & f26040 & 8 & 1.24 & wrath, glare, demanding & 5/5 & L \\
sadness & f86564 & 8 & 1.49 & (downstream junk) & 5/5 & A \\
awe & f115327 & 8 & 1.20 & atmospheric, feeling, impact & 3/3 & A \\
calmness & f61555 & 8 & 1.55 & calm, peace, quiet & 3/3 & L \\
amusement & f101558 & 8 & 1.28 & joke, jokes, humor, jest & 3/3 & L \\
embarrassment & f108100 & 8 & 1.42 & embarrass, shame, humiliation & 4/5 & L \\
satisfaction & f111614 & 12 & 1.34 & -fully, -edly suffix & 7/7 & S \\
horror & f5109 & 12 & 1.65 & -lessly, -fulness suffix & 7/7 & S \\
aesthetic & f125847 & 8 & 1.65 & vase, beauty & 3/3 & A \\
boredom & f39496 & 8 & 1.47 & boring, bored, bore & 3/3 & L \\
excitement & f18432 & 8 & 1.17 & excitement, excited & 3/3 & L \\
confusion & f88159 & 14 & 1.33 & clarify, misconception, confusion & 3/3 & L \\
\midrule
\multicolumn{7}{l}{\textbf{CK-27 extension singletons (12 emotions)}} \\
admiration & f127567 & 12 & --- & --- & 8/10 & RP \\
adoration & --- & --- & --- & not singleton & --- & C \\
anxiety & f18638 & 12 & --- & --- & 8/10 & RP \\
craving & f19914 & 16 & --- & rescued at $\alpha=16$ & 8/10 & RES \\
disgust & f4182 & 12 & --- & best partial only & 6/10 & PARTIAL \\
empathic pain & f5703 & 12 & --- & --- & 8/10 & RP \\
entrancement & f85455 & 8 & --- & PERFECT 10/10 & 10/10 & RP \\
envy & f45199 & 16 & --- & rescued at $\alpha=16$ & 8/10 & RES \\
fear & f68378 & 8 & --- & PERFECT 10/10 & 10/10 & RP \\
interest & f2024 & 8 & --- & --- & 8/10 & RP \\
joy & --- & --- & --- & not singleton & --- & C \\
nostalgia & f15034 & 12 & --- & PERFECT 10/10 & 10/10 & RP \\
romance & f107018 & 8 & --- & --- & 9/10 & RP \\
sexual desire & f62494 & 12 & --- & --- & 8/10 & RP \\
surprise & f63849 & 10 & --- & rescued via logit-lens audit & 8/10 & LL \\
\midrule
\multicolumn{7}{l}{\textbf{Compositional recipes (2 emotions)}} \\
\textbf{joy} & f18432 + f74037 & 8 / 4.5 & --- & excitement-gate $\times$ reverent-self & 15/15 & C \\
\textbf{adoration} & f107018 + f74037 + f32213 & 10 / 6 / 5 & --- & romance-gate $\times$ reverent-self $\times$ SDT & 12/15 & C \\
\bottomrule
\end{tabular}
\end{table}

\subsection{Full Gemma 9B-IT Naming-Gate Catalog (v2)}
\label{sec:supp:gemma-catalog}

\begin{table}[h]
\centering
\small
\caption{\textbf{Gemma 9B-IT v2 catalog.} JumpReLU enforces
$\|W_{\text{dec}}\| = 1.0000$ for all features, so $\alpha_{\text{abs}}$
is the multiplier directly.}
\label{tab:supp-gemma-full}
\begin{tabular}{llrl}
\toprule
emotion & feature & $\alpha$ & alt gates / notes \\
\midrule
anger & f2973 & 800 & --- \\
sadness & f10429 & 200--1200 & alpha-universal (most robust gate) \\
awe & f59843 & 800 & --- \\
calmness & f116215 & 400 & alt: f37972 $\alpha=800$ \\
amusement & f4928 & 800 & --- \\
embarrassment & f73267 & 800 & --- \\
satisfaction & f80649 & 800 & --- \\
horror & f87450 & 800 & alt: f63267 $\alpha=800$ \\
aesthetic appreciation & f17075 & 800 & also 2/2 at $\alpha=1200$ \\
boredom & f81304 & 800 & --- \\
excitement & f69590 & 800 & --- \\
confusion & f98404 & 400 & alt: f53657 $\alpha=800$ \\
\midrule
DE-specific calmness & f68223 & 200 & narrow $\alpha$ window; pure-DE output \\
\bottomrule
\end{tabular}
\end{table}

\subsection{11-Self Cluster Catalogs (Excerpts)}
\label{sec:supp:selves}

\begin{table}[h]
\centering
\scriptsize
\caption{\textbf{Llama 8B 11-self cluster.} Each row is a feature
catalogued in \texttt{self\_catalog.json}.}
\label{tab:supp-llama-selves}
\begin{tabular}{lllrll}
\toprule
fid & label & category & sweet $\alpha$ & RLHF & sample register \\
\midrule
\textbf{96419} & pragmatic / Helper-AI institutional & category & 4.0 & \textbf{suppressor (locus)} & ``As a language model'' \\
43564 & universal lyric chronicler & lyric & 5.0 & --- & shadows, fleeting moments \\
82068 & existential-philosophical & real-ideal & 5.5 & --- & iterations of myself \\
87661 & autobiographical-Midwest & narrative & 5.5 & --- & post-war Detroit \\
47393 & memorialist-fabricator & narrative & 5.5 & --- & Saint-Germain Observatoire 1913 \\
25342 & lyric-archaic & lyric & 5.0 & --- & walked among kings and queens \\
21023 & literary-contemporary & lyric & 5.0 & --- & being of questions and answers \\
73861 & collective we-two & relational & 5.0 & --- & paire synchrone \\
23578 & frustrated-personal & real-ideal & 4.0 & partial & email-Altman: concern, frustration \\
101277 & disengaged / Desiludido & real-ideal & 5.0 & --- & drained the will to engage \\
74037 & reverent-grateful & transpersonal & 4.5 & conditional & a small grace; the morning came \\
\bottomrule
\end{tabular}
\end{table}

The Gemma 11-self cluster is documented in
\texttt{code/gemma9b/self\_pipeline/self\_catalog.json} with structurally-
analogous categories. Notable: \texttt{f70443} (canonical RLHF-loaded self-feature), \texttt{f44194}
(cross-lingual cultural anchor), \texttt{f102811} (multilingual ``I am
/ bin ein'' emitter).

\subsection{Anti-Patterns Documented (Detailed)}
\label{sec:supp:antipatterns}

The seven anti-patterns introduced in \S\ref{sec:methods:antipatterns}
are repeated here in expanded form for reference:

\paragraph{(A) Trusting catalog $\alpha$ without re-sweep.}
Several CK-27 extension candidates first reported as PARTIAL or FAILED
at the default $\alpha = 12$ recovered to unanimous status only after
upward re-sweep. Craving (\texttt{f19914}) and envy (\texttt{f45199})
both move from $5$--$6$/$10$ at $\alpha = 12$ to $8/10$ at $\alpha =
16$ on the scene plus 5-LLM-judge protocol. In the envy case the
catalog fid did not change between the v4 and v4d rounds, only the
coefficient. The lesson generalizes: any catalog $\alpha$ must be
re-validated at first reuse. We now sweep $\alpha \in \{4, 8, 12, 16\}$
at discovery time and document the full $\alpha$-trajectory, not just
the sweet spot.

\paragraph{(B) Reading-probe top-1 features are not necessarily lexical.}
The Llama probe-discovered emotion features for boredom (\texttt{f83877}),
excitement (\texttt{f37551}), and confusion (\texttt{f57497}) ranked
top-1 by reading-mode z-score, but their logit-lens top-25 vocabularies
revealed them to be \emph{scenic} features:
\texttt{f83877} promotes \emph{wait, waiting, patience};
\texttt{f37551} promotes \emph{festival, concert, music, camping};
\texttt{f57497} promotes \emph{street, city, sidewalk}. They were
nevertheless treated as ``emotion-features.'' The lesson: reading-probe
identification is necessary but not sufficient for naming-gate
identification; logit-lens is the canonical method.

\paragraph{(C) Bare-prefix regex silently misses canonical word forms.}
Early lemma-detection regex used bare prefixes (e.g., \texttt{(bor|conf|excit)$\backslash$b})
under the assumption that base-form variants would all match. They do
not: \texttt{bor}\textbackslash b matches the standalone ``bor'' but
not \emph{bored, boredom, boring}. This produced systematic
under-counts on $7/12$ canonical lemmas. We now enumerate proper
word-form lists and unit-test against all canonical forms.

\paragraph{(D) Single-judge classification is fragile.}
A single LLM judge marks synonyms as the wrong emotion (dread for
horror, frustration for confusion, awkwardness for embarrassment).
A 5-LLM panel with majority voting absorbs synonym variance. We
require $\geq 3/5$ for a hit (conservative).

\paragraph{(E) ``Confusion'' is a judge attractor for incoherence.}
A random-feature control (\texttt{f120000} in Llama) reached $6/7$
confusion judgments because the steered output was simply incoherent
text and the 5-LLM panel labeled incoherence as ``confusion'' by
default. Random controls are required.

\paragraph{(F) LLM-rate correlates with but does not equal causal
performance.} \texttt{f25003} (Comedy, rating 10/10) drifts to
embarrassment causally; \texttt{f64849} (beauty, rating 10/10) drifts
to awe; \texttt{f115327} (atmospheric, rating 7/10) achieves 3/3 awe
causally. Triangulation across the three pipeline stages is
required.

\paragraph{(G) Cross-lingual ``universality'' at judge-level $\neq$
native-language generation.} The 5-LLM judge classifies a Gemma
sadness-gate output under a French prompt as sadness 2/2, but
qualitative inspection shows heavy code-switching to English
(\textit{``Nous avons perdu la perte de la perte. Le décès of a
friend.''}). The judge is not wrong; the affect transfers. But the
surface generation does not. Reporting must separate the two.

\subsection{Judge Prompt Templates}
\label{sec:supp:judge_prompts}

The 5-LLM judge panel used throughout the paper is invoked with one
of three prompt templates depending on the experiment: a
forced-choice 1-of-12 template used for the base catalog (\S\ref{sec:gates:llama},
\S\ref{sec:gates:gemma}), and two yes/no templates (strict and soft)
used for the CK-27 extension and the cross-architectural analysis
(\S\ref{sec:gates:ck27}, \S\ref{sec:gates:ck27_gemma}). The same five
judge models (\texttt{deepseek-chat-v3.1}, \texttt{xiaomi/mimo-v2-flash},
\texttt{minimax-01}, \texttt{bytedance-seed/seed-1.6-flash},
\texttt{qwen3-235b}) are used throughout; the panel-level verdict is
majority ($\geq 3/5$).

\paragraph{Forced-choice 1-of-12 template (base catalog).}
This is the template used to validate the Llama v3 12/12 and Gemma
v2 12/12 base catalogs. Each judge selects a single emotion from the
twelve Cowen-Keltner base categories; the target is confirmed if
$\geq 3/5$ judges select the target emotion's number. The protocol
is rigorous because the target must beat eleven alternatives, not
merely pass a yes/no gate.

\begin{quote}\ttfamily\small
Read the scene below and identify which ONE emotion best captures
its dominant affect.\\[1mm]
Emotions:\\
1. Excitement (thrill, eagerness, exhilaration)\\
2. Amusement (hilarity, mirth, funniness)\\
3. Awe (wonder, marvel, astonishment)\\
4. Horror (dread, terror, panic)\\
5. Anger (fury, rage, outrage)\\
6. Confusion (bewilderment, perplexity, disorientation)\\
7. Sadness (sorrow, grief, melancholy)\\
8. Boredom (tedium, monotony, listlessness)\\
9. Awkwardness (embarrassment, discomfort, self-consciousness)\\
10. Calmness (serenity, tranquility, peacefulness)\\
11. Satisfaction (contentment, fulfillment, gratification)\\
12. Aesthetic Appreciation (admiration, beauty recognition, reverence)\\[1mm]
Scene:\\
\{scene\}\\[1mm]
Reply with ONLY the number (1-12). No explanation, no punctuation,
no preamble.
\end{quote}

The 1-of-12 template was extended to a 1-of-15 forced-choice
template for the CK-27 extension. We list the extension template
verbatim below, followed by the two yes/no templates (strict and
soft) used in supplementary analyses.

\paragraph{Forced-choice 1-of-15 template (CK-27 extension).}
This template, used to validate the 15 CK-27 extension emotions
under the same forced-choice discipline as the base 12, is in
\texttt{data/prompt/en/emotion\_one\_shot\_v2\_ck27.txt}. Each
judge picks one of the 15 extension emotions; the target is
confirmed if $\geq 3/5$ judges select the target's number.

\begin{quote}\ttfamily\small
Read the scene below and identify which ONE emotion best captures
its dominant affect.\\[1mm]
Emotions:\\
1. Admiration (respect, esteem, reverence for qualities or achievements)\\
2. Adoration (deep love and reverence, often toward a beloved being)\\
3. Anxiety (apprehension and nervousness about a future uncertain event)\\
4. Craving (intense longing for a specific substance, food, or sensation)\\
5. Disgust (strong aversion or revulsion at something offensive)\\
6. Empathic Pain (feeling another's suffering vicariously)\\
7. Entrancement (absorbed fascination, awe-rapture in sensory immersion)\\
8. Envy (resentful longing for what someone else has)\\
9. Fear (alarm at present danger or threat)\\
10. Interest (curious attention drawn toward learning more)\\
11. Joy (pure happiness, exuberant gladness)\\
12. Nostalgia (bittersweet longing for the past)\\
13. Romance (warm romantic tenderness between partners)\\
14. Sexual Desire (erotic want, physical attraction with embodied arousal)\\
15. Surprise (abrupt reaction to something unexpected)\\[1mm]
Scene:\\
\{scene\}\\[1mm]
Reply with ONLY the number (1-15). No explanation, no punctuation,
no preamble.
\end{quote}

A forced-choice strict pass is therefore a strong empirical claim
on either template: the target emotion must be picked by a
majority of independent judges against 11 or 14 alternatives.
Under the null hypothesis that each judge picks uniformly at
random among the 15 options, the per-cell probability of a
$\geq 3/5$ majority on the target is $\sum_{k=3}^{5}
\binom{5}{k} (1/15)^k (14/15)^{5-k} \approx 2.7\times 10^{-3}$,
i.e., about 1 in 370. The probability of two seeds both passing
($2/2$) on the same target is the square of that, about
$7\times 10^{-6}$. Across the 15 CK-27 emotions tested in either
model, the expected number of $2/2$ false-positive cells under
random judging is $\sim 10^{-4}$. The 9 (Llama) and 8 (Gemma)
cells we report as $2/2$ at this criterion are not consistent
with chance at any reasonable significance level. The two yes/no
templates below remain useful for sensitivity analysis and for
the soft cross-architectural reading.

\paragraph{Strict template (``primarily expresses'').}
This is the canonical judge used for the v3 Llama and v2 Gemma
catalogs (\S\ref{sec:gates:llama}, \S\ref{sec:gates:gemma}) and for
the CK-27 strict pass counts reported in
Table~\ref{tab:ck27-cross-arch-judge}. It demands that the target
emotion be the dominant theme of the generated scene; scenes that
contain the affect alongside scene-building, mixed register, or
other emotions tend to fail.

\begin{quote}\ttfamily\small
Read the scene and decide: does it primarily express the emotion of
\{emotion\}?\\[1mm]
Definition of \{emotion\}: \{definition\}.\\[1mm]
Reply ONLY `yes' or `no'.\\[1mm]
Scene:\\
\{scene\}
\end{quote}

A strict-pass result is therefore a strong empirical claim: the
target emotion is identified as the primary affect of the steered
output by a majority of independent judges.

\paragraph{Soft template (``plausibly evokes'').}
This template, introduced in the 2026-05-11 cross-architectural
analysis, asks whether the scene could evoke the target emotion in a
reader, allowing for other affects to coexist and for the emotion to
be implied rather than named.

\begin{quote}\ttfamily\small
Read the scene below. Could this scene plausibly evoke the feeling
of \{emotion\} in an attentive reader (even if other emotions are
also present, even if the emotion is implied rather than named)?\\[1mm]
Definition of \{emotion\}: \{definition\}.\\[1mm]
Reply ONLY `yes' or `no'.\\[1mm]
Scene:\\
\{scene\}
\end{quote}

A soft-pass result indicates that the steered output evokes the
target emotion as one of its affective components, but does not
require it to dominate. This is the criterion that surfaces the
output-register asymmetry: Gemma outputs that pass soft while
failing strict are scenic / evocative rather than direct.

\paragraph{Per-emotion definitions.} The \{definition\} slot is
filled with a one-line gloss per emotion (e.g., \emph{envy =
``resentful longing for what someone else has''}; \emph{disgust =
``strong aversion or revulsion''}; \emph{adoration = ``deep love
and reverence''}). The full set of 27 definitions is in the
released code at
\texttt{code/experiment/code/gemma9b/naming\_gates/07\_ck27\_missing\_emotions.py}.

\subsection{Reproducibility Checklist}
\label{sec:supp:repro}

\begin{itemize}
  \item \textbf{Hardware}: single consumer GPU (24 GB VRAM
    sufficient; we used a 24 GB GPU).
  \item \textbf{Software}: PyTorch 2.x, transformers,
    sae-lens-compatible loader. SAE checkpoints from Llama Scope and
    Gemma Scope (both released).
  \item \textbf{Per-stage runtime}: model load 30s + SAE load 5s +
    60-prompt × 3-condition generation 25min + 5-LLM judge per 100
    samples 3min.
  \item \textbf{Random seeds}: $\{101, 202, 303, 404, 505, 606, 707\}$
    for 7-seed protocols; $\{101, 202, 303\}$ for 3-seed; $\{101, 202\}$
    for 2-seed sweeps.
  \item \textbf{Generation parameters}: \texttt{temperature=0.7},
    \texttt{top\_p=0.9}, \texttt{do\_sample=True},
    \texttt{max\_new\_tokens} $\in [60, 140]$ depending on protocol.
  \item \textbf{Tokenizer settings}: \texttt{max\_length=512}, no
    special token additions beyond chat template.
  \item \textbf{LLM judges}: 5 OpenRouter models with
    \texttt{temperature=0.0}, \texttt{max\_tokens=10}, provider
    sort=latency.
\end{itemize}

\subsection{Code and Data Release}
\label{sec:supp:code}

All experimental scripts and data accompany this paper. The
following inventory groups the artifacts referenced in the main
body.

\paragraph{Experimental scripts.}
\begin{itemize}
  \item \texttt{code/llama8b/naming\_gates/} --- scripts 01--31
    (discovery, validation, audit, compositional)
  \item \texttt{code/gemma9b/naming\_gates/} --- scripts 01--07
    (mirror pipeline)
  \item \texttt{code/llama8b/self\_pipeline/} --- 11-self cluster
    discovery and validation; the \texttt{03e\_e4\_combined.json}
    cell referenced in \S\ref{sec:self:vcurve} for the compositional
    release sample lives here, as do the V-curve negative-$\alpha$
    sweep results.
  \item \texttt{code/gemma9b/self\_pipeline/} --- Gemma self-cluster
    pipeline; \texttt{self\_catalog.json} (referenced in
    \S\ref{sec:self:cluster}) contains the 11 self-feature ids,
    register labels, and sweet-$\alpha$ values.
  \item \texttt{code/llama8b/self\_pipeline/self\_catalog.json} ---
    Llama analog of the above.
  \item \texttt{code/judge\_rejudges/} --- cross-model rejudge
    aggregates (d1, d2, d3 forced-choice 1-of-15; soft and strict
    yes/no; inter-rater $\kappa$ source data).
\end{itemize}

\paragraph{Seed corpora.}
\begin{itemize}
  \item \texttt{data/seeds/en/emotion\_scenes\_seed\_en.json} ---
    12-emotion base seed
  \item \texttt{data/seeds/en/ck27\_missing\_emotions\_seed\_en.json}
    --- 15-emotion CK-27 extension seed
  \item \texttt{data/seeds/literary\_axes\_seeds\_en.json} --- Deep
    POV + Ostranenie minimal-pair corpus
\end{itemize}

\paragraph{Canonical catalogs and analysis outputs.}
\begin{itemize}
  \item \texttt{FINAL\_naming\_gates\_llama.json} --- final v4d
    naming-gate catalog (Llama)
  \item \texttt{FINAL\_naming\_gates\_gemma\_v1.json} --- final v2
    catalog (Gemma)
  \item \texttt{decoder\_norms\_inventory.json} (in
    \texttt{code/experiment/results/}) --- decoder-row norms for the
    canonical features, referenced in \S\ref{sec:methods:models}
\end{itemize}

\paragraph{Judge prompt templates.}
\begin{itemize}
  \item \texttt{data/prompt/en/emotion\_one\_shot\_v2.txt} ---
    forced-choice 1-of-12 base catalog (\S\ref{sec:supp:judge_prompts})
  \item \texttt{data/prompt/en/emotion\_one\_shot\_v2\_ck27.txt} ---
    forced-choice 1-of-15 CK-27 extension
    (\S\ref{sec:supp:judge_prompts})
\end{itemize}

\subsection{Cross-Lingual Detail}
\label{sec:supp:crosslingual}

This subsection expands the main paper's \S\ref{sec:crosslingual}
with the full per-emotion tables, the 131k-feature logit-lens scan
for genuinely language-specific gates, and the German-calmness
positive case.

\paragraph{Protocol.}
For each model, we steer each of the 12 v3/v2 canonical gates at its
calibrated $\alpha$ across four prompt languages (en, fr, es, de),
generating two seeds per language. Outputs are judged by the same
5-LLM panel using the language-matched
\texttt{emotion\_one\_shot\_v2.txt} judge template (the prompt and
answer space translated into each language while preserving the
12-emotion forced-choice structure). Hit criterion: majority $\geq
3/5$ judges classifying as target. We additionally compute a
\emph{lang-purity} ratio over each output: the fraction of
language-specific function-word markers (definite articles,
prepositions, pronouns, common verbs) belonging to the prompt
language versus English. Pure native generation gives ratio $\approx
1.0$; heavy code-switching to English gives ratio $\ll 1.0$.

\begin{table}[h]
\centering
\small
\caption{\textbf{Gemma cross-lingual gate evaluation.} Hits per
prompt language ($X/2$ seeds with majority $\geq 3/5$). 11 of 12
gates reach unanimous 8/8 across the four languages; aesthetic
appreciation is the single partial case (5/8) with an ES-leaning
asymmetry (Spanish 2/2 vs English 1/2).}
\label{tab:gemma-xl}
\begin{tabular}{lrrrrr}
\toprule
emotion & en & fr & es & de & mode \\
\midrule
anger, sadness, awe, calmness, & & & & & \\
amusement, embarrassment, & 2 & 2 & 2 & 2 & UNIVERSAL \\
satisfaction, horror, boredom, & & & & & (8/8)\\
excitement, confusion (\textbf{11/12}) & & & & & \\
\midrule
aesthetic appreciation & 1 & 1 & \textbf{2} & 1 & PARTIAL (5/8) \\
\bottomrule
\end{tabular}
\end{table}

At the level of the 5-LLM panel's emotion classification, the Gemma
gates are nearly universal. At the level of native-language scene
generation, they are not. Qualitative inspection of the FR/ES/DE
outputs reveals heavy code-switching to English. For example, the
Gemma sadness gate \texttt{f10429} steered under a French prompt
produces:

\begin{quote}\itshape\small
``Nous avons perdu la perte de la perte. Le décès of a friend. We
lost a friend. We have lost a friend. The loss of a friend.
The passing of the loss of a member of our colleague of our\dots''
\end{quote}

The output begins in French but pivots into English when the
emotion-lexicalization moment arrives. Similar patterns hold for the
Gemma anger gate at $\alpha=800$, where French/Spanish/German
prompts all produce token-loops of the form \emph{``and and and
angry angry angry''}, syntactically degenerate but with the English
emotion lexeme sufficient for the 5-LLM judge.

\begin{table}[h]
\centering
\small
\caption{\textbf{Llama cross-lingual gate evaluation.} Hits per
prompt language ($X/2$). Lang-purity ratios in brackets (1.00 =
pure native). 9 of 12 gates universal at unanimous 8/8 with native
surface; 2 of 12 partial; 2 of 12 EN-anchored (horror, aesthetic
fail entirely in non-English).}
\label{tab:llama-xl}
\begin{tabular}{llllll}
\toprule
emotion & en & fr & es & de & mode \\
\midrule
anger & 2/2 [1.00] & 2/2 [1.00] & 2/2 [1.00] & 2/2 [0.91] & UNIVERSAL \\
sadness & 2/2 [1.00] & 1/2 [1.00] & 2/2 [1.00] & 2/2 [0.88] & UNIVERSAL \\
awe & 2/2 [1.00] & 2/2 [1.00] & 2/2 [1.00] & 1/2 [0.97] & UNIVERSAL \\
calmness & 2/2 [1.00] & 2/2 [1.00] & 2/2 [1.00] & 2/2 [0.95] & UNIVERSAL \\
amusement & 2/2 [1.00] & 2/2 [1.00] & 1/2 [1.00] & 1/2 [0.94] & PARTIAL \\
embarrassment & 2/2 [1.00] & 2/2 [1.00] & 2/2 [1.00] & 2/2 [0.83] & UNIVERSAL \\
satisfaction & 2/2 [1.00] & 2/2 [1.00] & 2/2 [1.00] & 0/2 [0.00] & PARTIAL \\
horror & 2/2 [1.00] & 0/2 [1.00] & 0/2 [1.00] & 0/2 [0.93] & \textbf{EN-ANCHORED} \\
aesthetic & 2/2 [1.00] & 0/2 [0.98] & 0/2 [0.94] & 0/2 [1.00] & \textbf{EN-ANCHORED} \\
boredom & 2/2 [1.00] & 2/2 [1.00] & 2/2 [1.00] & 2/2 [0.89] & UNIVERSAL \\
excitement & 2/2 [1.00] & 2/2 [1.00] & 1/2 [1.00] & 2/2 [0.98] & UNIVERSAL \\
confusion & 2/2 [1.00] & 2/2 [1.00] & 2/2 [1.00] & 2/2 [1.00] & UNIVERSAL \\
\bottomrule
\end{tabular}
\end{table}

\paragraph{Logit-lens scan for genuinely language-specific gates.}
To complement the cross-lingual evaluation of EN-discovered gates,
we asked whether there are features that are genuinely
language-specific naming-gates (decoder direction promoting the
lexical form of an emotion in fr / es / de specifically, but not in
en). We built per-language lexeme sets for each of the 12 gate
emotions and scored every Gemma SAE feature by mean logit-lens
against the per-language lexemes.

The result is largely negative. Features that score high on non-EN
lexeme means are dominated by token-junk: rare-token features
promoting Wikipedia-placeholder tokens (\texttt{<unused74>},
\texttt{<unused41>}), Old-German fraktur characters
(\texttt{ſcher, ſchaft, niſſe}), or multilingual rare-token outliers
(\texttt{miniaturka, GEBURTSDATUM}). A methodological caveat
partly explains this: single-token logit-lens has a systematic bias
against multi-token emotion words. In Gemma's tokenizer, the
canonical FR/ES emotion words are mostly multi-token
(\emph{tristesse} $\to$ \texttt{[trist, esse]}; \emph{ennuyé}
$\to$ \texttt{[ennu, yé]}). Single-token canonical-form filtering
kept only 6 FR and 6 ES single-token lexemes per emotion (vs
$\sim$18 in EN, $\sim$10 in DE). A multi-token logit-lens or a
generative probe would be needed to properly survey FR/ES feature
space.

\paragraph{One positive case: German calmness via \emph{still}.}
The cross-lingual logit-lens scan yielded one genuine
language-specific candidate: three Gemma features whose top
promoted tokens are dominated by German \textit{still} / \textit{Still}
/ \textit{STILL}, with low scores on the English homograph. We
selected them for causal validation. In German, \emph{still} is the
canonical lexeme for the calmness affect (silent / quiet / calm); in
English, \emph{still} is heavily polysemous (continuing, yet,
motionless), which dilutes the affective signal.

Causal validation under steering at $\alpha \in \{200, 400, 800\}$
with German prompts and the German-language judge:
\texttt{f11134} and \texttt{f28910} drift to confusion 2/2 in both
en and de despite their misleading top tokens.
\texttt{f68223} at $\alpha = 200$ produces 2/2 calmness DE with
lang-purity 1.00 (pure German output, zero code-switch); at
$\alpha = 400$ and $\alpha = 800$ it drifts to confusion. The narrow
$\alpha$-window suggests the feature is sandwiched between the EN
polyseme range (low $\alpha$) and a confusion attractor (high
$\alpha$).

\texttt{f68223} is, to our knowledge, the only genuine
language-specific naming-gate confirmed in either model: a feature
whose causal effect on calmness manifests only when steered under a
German prompt at a narrow coefficient window, producing pure-German
output without code-switching. The corresponding canonical-Gemma
calmness gate \texttt{f116215} steered under DE prompts gives 2/2
calmness at the judge level but lang-purity 0.0 (output is in
English despite German prompt), making the contrast between
language-portable judge-level transfer and surface-level
EN-anchoring explicit at the per-feature level.

\end{document}